\documentclass[journal]{IEEEtran}
\usepackage{amsmath,amssymb,amsfonts}
\usepackage[linesnumbered,ruled,vlined]{algorithm2e}
\usepackage{pgfplots}
\usepackage{tikz}
\usepackage{tkz-tab}

\usepackage{tabularx}
\usepackage{float}
\usetikzlibrary{patterns}
\usepgfplotslibrary{statistics}
\pgfplotsset{compat=1.18}
\pgfmathdeclarefunction{fpumod}{2}{%
    \pgfmathfloatdivide{#1}{#2}%
    \pgfmathfloatint{\pgfmathresult}%
    \pgfmathfloatmultiply{\pgfmathresult}{#2}%
    \pgfmathfloatsubtract{#1}{\pgfmathresult}%
    \pgfmathfloatifapproxequalrel{\pgfmathresult}{#2}{\def\pgfmathresult{5}}{}%
}
\pgfplotsset{boxplot legend/.style={
    legend image code/.code={
        \draw[#1] (0cm,-0.1cm) rectangle (0.4cm,0.1cm)
        (0.2cm,-0.1cm) -- (0.2cm,-0.2cm) (0.05cm,-0.2cm) -- (0.35cm,-0.2cm)
        (0.2cm,0.1cm) -- (0.2cm,0.2cm) (0.05cm,0.2cm) -- (0.35cm,0.2cm);
     \path (0cm,0.24cm) (0cm,-0.24cm);  
    },
}}

\usepackage{amsmath,amsfonts}
\usepackage{bm}
\usepackage[colorlinks = true,
linkcolor = blue,
urlcolor  = blue,
citecolor = blue,
anchorcolor = blue]{hyperref}
\usepackage{tikz}
\usetikzlibrary{patterns}
\usepackage{array}
\usepackage{breqn}
\usepackage{textcomp}
\usepackage{stfloats}
\usepackage{url}
\usepackage{pgfplots}
\usepackage{verbatim}
\usepackage{tikz}
\usepackage{subcaption}
\usepackage{graphicx}
\usepackage{amssymb}
\usepackage{multirow}
\usepackage{caption}
\usepackage{booktabs}
\usepackage{cite}
\usepackage{amsmath}
\pgfplotsset{compat=1.18} 

\usepackage[linesnumbered,ruled,vlined]{algorithm2e}
\usepackage{rotating}
\usepackage[section]{placeins}

\usepackage[linesnumbered,ruled,vlined]{algorithm2e}

\usepackage{multicol}

\usetikzlibrary{patterns}
\usepackage{xcolor}
\usetikzlibrary{calc}
\usepackage{pgfplots}
\usepackage{pgfplotstable}
\pgfplotsset{compat=1.16}
\pgfkeys{/tikz/.cd,
cube top color/.store in=\CubeTopColor,
cube top color=blue!60,
cube front color/.store in=\CubeFrontColor,
cube front color=blue!30,
cube side color/.store in=\CubeSideColor,
cube side color=blue!40,
3d cube color/.code={\colorlet{mycolor}{#1}%
	\tikzset{cube top color=mycolor!60,cube front color=mycolor!30,%
		cube side color=mycolor!40,draw=mycolor}}
		}
		\makeatletter
		\pgfdeclareplotmark{half cube*}
		{%
\pgfplots@cube@gethalf@x
\let\pgfplots@cube@halfx=\pgfmathresult
\pgfplots@cube@gethalf@y
\let\pgfplots@cube@halfy=\pgfmathresult
\pgfplots@cube@gethalf@z
\let\pgfplots@cube@halfz=\pgfmathresult
\pgfmathparse{0*\pgfplots@cube@halfz}%
\let\pgfplots@cube@topz=\pgfmathresult
\pgfmathparse{-1*\pgfplots@cube@halfz}%
\let\pgfplots@cube@bottomz=\pgfmathresult
\pgfplotsifaxissurfaceisforeground{0vv}{%
	\pgfsetfillcolor{\CubeFrontColor}
	\pgfpathmoveto{\pgfplotsqpointxyz{-\pgfplots@cube@halfx}{-\pgfplots@cube@halfy}{\pgfplots@cube@bottomz}}%
	\pgfpathlineto{\pgfplotsqpointxyz{-\pgfplots@cube@halfx}{-\pgfplots@cube@halfy}{\pgfplots@cube@topz}}%
	\pgfpathlineto{\pgfplotsqpointxyz{-\pgfplots@cube@halfx}{ \pgfplots@cube@halfy}{\pgfplots@cube@topz}}%
	\pgfpathlineto{\pgfplotsqpointxyz{-\pgfplots@cube@halfx}{ \pgfplots@cube@halfy}{\pgfplots@cube@bottomz}}%
	\pgfpathclose
	\pgfusepathqfillstroke
}{%
	\pgfsetfillcolor{\CubeFrontColor}
	\pgfpathmoveto{\pgfplotsqpointxyz{ \pgfplots@cube@halfx}{-\pgfplots@cube@halfy}{\pgfplots@cube@bottomz}}%
	\pgfpathlineto{\pgfplotsqpointxyz{ \pgfplots@cube@halfx}{-\pgfplots@cube@halfy}{\pgfplots@cube@topz}}%
	\pgfpathlineto{\pgfplotsqpointxyz{ \pgfplots@cube@halfx}{ \pgfplots@cube@halfy}{\pgfplots@cube@topz}}%
	\pgfpathlineto{\pgfplotsqpointxyz{ \pgfplots@cube@halfx}{ \pgfplots@cube@halfy}{\pgfplots@cube@bottomz}}%
	\pgfpathclose
	\pgfusepathqfillstroke
}%
\pgfplotsifaxissurfaceisforeground{v0v}{%
	\pgfsetfillcolor{\CubeSideColor}
	\pgfpathmoveto{\pgfplotsqpointxyz{-\pgfplots@cube@halfx}{-\pgfplots@cube@halfy}{\pgfplots@cube@bottomz}}%
	\pgfpathlineto{\pgfplotsqpointxyz{-\pgfplots@cube@halfx}{-\pgfplots@cube@halfy}{\pgfplots@cube@topz}}%
	\pgfpathlineto{\pgfplotsqpointxyz{ \pgfplots@cube@halfx}{-\pgfplots@cube@halfy}{\pgfplots@cube@topz}}%
	\pgfpathlineto{\pgfplotsqpointxyz{ \pgfplots@cube@halfx}{-\pgfplots@cube@halfy}{\pgfplots@cube@bottomz}}%
	\pgfpathclose
	\pgfusepathqfillstroke
}{%
	\pgfsetfillcolor{\CubeSideColor}
	\pgfpathmoveto{\pgfplotsqpointxyz{-\pgfplots@cube@halfx}{ \pgfplots@cube@halfy}{\pgfplots@cube@bottomz}}%
	\pgfpathlineto{\pgfplotsqpointxyz{-\pgfplots@cube@halfx}{ \pgfplots@cube@halfy}{\pgfplots@cube@topz}}%
	\pgfpathlineto{\pgfplotsqpointxyz{ \pgfplots@cube@halfx}{ \pgfplots@cube@halfy}{\pgfplots@cube@topz}}%
	\pgfpathlineto{\pgfplotsqpointxyz{ \pgfplots@cube@halfx}{ \pgfplots@cube@halfy}{\pgfplots@cube@bottomz}}%
	\pgfpathclose
	\pgfusepathqfillstroke
}%
\pgfplotsifaxissurfaceisforeground{vv0}{%
	\pgfsetfillcolor{\CubeTopColor}
	\pgfpathmoveto{\pgfplotsqpointxyz{-\pgfplots@cube@halfx}{-\pgfplots@cube@halfy}{\pgfplots@cube@bottomz}}%
	\pgfpathlineto{\pgfplotsqpointxyz{-\pgfplots@cube@halfx}{ \pgfplots@cube@halfy}{\pgfplots@cube@bottomz}}%
	\pgfpathlineto{\pgfplotsqpointxyz{ \pgfplots@cube@halfx}{ \pgfplots@cube@halfy}{\pgfplots@cube@bottomz}}%
	\pgfpathlineto{\pgfplotsqpointxyz{ \pgfplots@cube@halfx}{-\pgfplots@cube@halfy}{\pgfplots@cube@bottomz}}%
	\pgfpathclose
	\pgfusepathqfillstroke
}{%
	\pgfsetfillcolor{\CubeTopColor}
	\pgfpathmoveto{\pgfplotsqpointxyz{-\pgfplots@cube@halfx}{-\pgfplots@cube@halfy}{\pgfplots@cube@topz}}%
	\pgfpathlineto{\pgfplotsqpointxyz{-\pgfplots@cube@halfx}{ \pgfplots@cube@halfy}{\pgfplots@cube@topz}}%
	\pgfpathlineto{\pgfplotsqpointxyz{ \pgfplots@cube@halfx}{ \pgfplots@cube@halfy}{\pgfplots@cube@topz}}%
	\pgfpathlineto{\pgfplotsqpointxyz{ \pgfplots@cube@halfx}{-\pgfplots@cube@halfy}{\pgfplots@cube@topz}}%
	\pgfpathclose
	\pgfusepathqfillstroke
}%
}
\makeatother

\begin{document}
\newcommand{\qnn}{QuantoN-SO}
\title{A Comprehensively Adaptive Architectural Optimization-Ingrained Quantum Neural Network Model for Cloud Workloads Prediction}
\author{Jitendra Kumar,~\IEEEmembership{Senior member,~IEEE,} Deepika Saxena,~\IEEEmembership{Member,~IEEE,} Kishu Gupta,~\IEEEmembership{Member,~IEEE,} \\ Satyam Kumar, Ashutosh Kumar Singh,~\IEEEmembership{Senior member,~IEEE}
	\thanks{Received 14 July 2024; revised 17 February 2025 and 24 May 2025; accepted 4 June 2025. This work was supported in part by the JSPS KAKENHI Early Career Research under Grant H-2024-13; in part by the University of Aizu, Japan, under CRF Grant FY2025; in part by Maulana Azad National Institute of Technology, Bhopal, India, Indian Institute of Information Technology, Bhopal, Madhya Pradesh, India; and in part by the National Sun Yat-sen University, Kaohsiung, Taiwan. (Corresponding author: Deepika Saxena.)}
	\thanks{Jitendra Kumar is with Department of Mathematics, Bioinformatics and Computer Applications Maulana Azad National Institute of Technology Bhopal, India. (email: jitendrakumar@ieee.org).\\
		Deepika Saxena, is with the School of Computer Science and Engineering, The University of Aizu, Aizuwakamatsu 965-0006, Japan and also with Department of Computer Science, the VIZJA University, 01-043 Warsaw, Poland. (e-mail: deepika@u-aizu.ac.jp). \\
		Kishu Gupta, is  with the Department of Computer Science and Engineering, National Sun Yat-sen University, Kaohsiung, 804, Taiwan. (e-mail: kishuguptares@gmail.com).\\
		Satyam Kumar, is with Department of Computer Applications, National Institute of Technology Tiruchirappalli, India. (e-mail: satyamk618@gmail.com). \\
		Ashutosh Kumar Singh, is with Department of Computer Science and Engineering, Indian Institute of Information Technology Bhopal, Bhopal 462003, India, and also with Department of Computer Science, the VIZJA University, 01-043 Warsaw, Poland. (e-mail: ashutosh@iiitbhopal.ac.in).}
\thanks{Digital Object Identifier 10.1109/TNNLS.2025.3577721}}


 \markboth{IEEE TRANSACTIONS ON NEURAL NETWORKS AND LEARNING SYSTEMS}%
{Shell \MakeLowercase{\textit{Kumar et al.}}: A Sample Article Using IEEEtran.cls for IEEE Journals}

\makeatletter
\newcommand{\removelatexerror}{\let\@latex@error\@gobble}
\def\ps@IEEEtitlepagestyle{%
	\def\@oddfoot{\mycopyrightnotice}%
	\def\@oddhead{\hbox{}\@IEEEheaderstyle\leftmark\hfil\thepage}\relax
	\def\@evenhead{\@IEEEheaderstyle\thepage\hfil\leftmark\hbox{}}\relax
	\def\@evenfoot{}%
}

\def\mycopyrightnotice{%
	\begin{minipage}{\textwidth}
		\centering \scriptsize
		2162-237X©2025 IEEE. All rights reserved, including rights for text and data mining, and training of artificial intelligence and similar technologies. Personal use is permitted, but republication/redistribution requires IEEE permission. This article has been accepted in IEEE Transactions on NEURAL NETWORKS AND LEARNING SYSTEMS © 2025 IEEE. This article has been accepted for inclusion in a future issue of this journal. Content is final as presented, with the exception of pagination. \\
		See https://www.ieee.org/publications/rights/index.html for more information. 
	\end{minipage}
}
\makeatother

\maketitle

\begin{abstract} 
Accurate workload prediction and advanced resource reservation are indispensably crucial for managing dynamic cloud services. Traditional neural networks and deep learning models frequently encounter challenges with diverse, high-dimensional workloads, especially during sudden resource demand changes, leading to inefficiencies. This issue arises from their limited optimization during training, relying only on parametric (inter-connection weights) adjustments using conventional algorithms. To address this issue, this work proposes a novel \textit{C}omprehensively \textit{A}daptive Architectural Optimization-based Variable \textit{Q}uantum \textit{N}eural \textit{N}etwork \textit{(CA-QNN)}, which combines the efficiency of quantum computing with complete structural and qubit vector parametric learning. The model converts workload data into qubits, processed through qubit neurons with Controlled NOT-gated activation functions for intuitive pattern recognition. In addition, a comprehensive architecture optimization algorithm for networks is introduced to facilitate the learning and propagation of the structure and parametric values in variable-sized QNNs. This algorithm incorporates quantum adaptive modulation and size-adaptive recombination during training process. The  performance of CA-QNN model is thoroughly investigated against seven state-of-the-art methods across four benchmark datasets of heterogeneous cloud workloads. The proposed model demonstrates superior prediction accuracy, reducing prediction errors by up to 93.40\% and 91.27\% compared to existing  deep learning and QNN-based approaches. 
\end{abstract}



\IEEEpeerreviewmaketitle

\section*{Nomenclature}
				$V$: VM \\
				$S$: servers \\ 
				$P$: \#servers \\
				$Q$: \#VMs \\ 
				$\alpha, \beta$: recombination points \\
				 $m^{\ast}$: \#raw data samples \\ 
				$m$: \#training input data samples \\  
				$Pn$: population \\ 
				$D_i$: $i^{th}$data sample \\ 
				$\widehat{D_i}$: $i^{th}$normalized data samples\\ 
				$\mathcal{Q}^{CN}$: C-NOT gate \\ $\widehat{{Fore}(D)_{i}}$: $i^{th}$forecasted normalized data value \\ 
				$\#$: training data samples \\ $n, q$: number of nodes in input and output layers \\ 
				$p^{\ast}$: variable number of hidden nodes \\ $K$: number of hidden layers  \\ 
				$VQNN$: variable QNN \\ $N$: size of population of variable network vectors    \\ 
				$\digamma$: fitness score \\  $Y^i$: intermediate qubit vector at $i^{th}$layer \\ 
				$I$: input layer \\  $H$: hidden layer\\ 
				$O$: output layer \\  $\Psi_i$: $i^{th}$input qubit data value  \\ 
				$\mathcal{Q}^R$: rotation gate \\ $B$: bias \\ 
				$\mathcal{S}(\varrho)$: sigmoid function \\
				$U_j$: magnitude of $j^{th}$ qubit vector argument \\ 
				$\uplus$: network qubit vector \\ $\Delta$: modulated qubit vector \\ 
				$R^{mod}$: modulation rate \\  $R$:random values \\ 
				$T_1, T_2, T_3$: modulation probabilities \\   $L$: size of VQNN vector \\ 
				$M^{\ast}$: selected modulation strategy \\ $\tau_1, \tau_2, \tau_3$: number of failed candidates \\ 
				$\varepsilon_{1}, \varepsilon_{2}, \varepsilon_{3}$: number of successful candidates \\ $C^j$: $j^{th}$ offspring vector \\ \\ \\
				
\section{Introduction}

\IEEEPARstart{C}{loud computing} has transformed information technology by offering diverse services over the Internet, including networking, storage, databases, servers, and software. These services have been widely adopted by individuals and enterprises to meet various needs. Large-scale data centers, established by cloud service providers, form the backbone of the cloud computing, enabling on-demand services for users \cite{tuli2021start, saxena2023performance}. With the growing reliance on cloud services, efficient workload management is crucial to maintain service quality, optimize resource allocation, and reduce operational costs \cite{kim2020forecasting}. However, the dynamic fluctuations in cloud workloads and resource usage can compromise service quality, degrade performance, and lead to outages, resource mismanagement, and excessive energy consumption \cite{yuan2016ttsa,  peng2020throughput}. Accurately predicting cloud workload demand is essential for effective resource provisioning and management, ensuring adaptive allocation and addressing these challenges \cite{feng2022fast,  saxena2022high}. Thus, accurate prediction of heterogeneous cloud workloads is vital for managing the entire cloud infrastructure efficiently.

\subsection{Related Work} \label{rw}

Accurate cloud workload forecasting has been a critical research area, with numerous conventional and advanced neural network-based models proposed to enhance predictive accuracy and computational efficiency. Kumar et al. \cite{kumar2018workload} introduced a dynamic cloud workload prediction model using an artificial neural network optimized via a self-adaptive differential evolution (SaDE) algorithm. By integrating multi-dimensional exploration and convergence strategies, this approach significantly improved predictive accuracy over traditional Backpropagation-trained networks \cite{lu2016rvlbpnn}. Building upon this, Kumar et al. \cite{kumar2020biphase} developed a Biphase-adaptive Differential Evolution (BaDE) learning-based neural network, further refining workload forecasting by dynamically balancing exploration and exploitation for optimized neural weight adjustments.

To address the challenges of capturing long-term dependencies in workload prediction, Kumar et al. \cite{kumar2018long} proposed a fine-grained model employing long short-term memory recurrent neural networks (LSTM-RNN). While this approach effectively captured temporal dependencies and improved host load predictions, its computational cost remained high due to the iterative Backpropagation (BPNN) process across recurrent layers. In contrast, the CloudInsight (CLIN) model \cite{kim2020forecasting} tackled resource misallocation issues caused by dynamic workload fluctuations by leveraging an ensemble of multiple predictors, dynamically optimized in real-time to outperform conventional models.

Expanding the scope beyond classical neural architectures, Li et al. \cite{li2021reinforcement} introduced a reinforcement learning-based online resource partitioning framework (RL-RP) for handling dynamic workload variations in cloud gaming environments. This approach demonstrated improved adaptability, outperforming traditional static partitioning methods. More recently, quantum-inspired models have gained traction, leveraging quantum computing principles to enhance workload forecasting.  Kumar et al. \cite{kumar2023quantum} proposed a Quantum Controlled-NOT Gate-based feed-forward neural network (QC-FNN), optimized with a Quantum Adaptive Differential Evolutionary learning algorithm, significantly improving precision in regional load forecasting. Similarly, Singh et al. \cite{singh2021quantum} introduced an Evolutionary Quantum Neural Network (EQNN), which exploited qubit diversity and quantum computation efficiency to enhance prediction accuracy.

Further advancements in architecture search and optimization methodologies have refined predictive efficiency. Zhang et al. \cite{zhang2023enhanced} developed a two-stage Neural Architecture Search (NAS) framework, integrating an enhanced gradient-based method and an evolutionary multi-objective algorithm to optimize network complexity and performance. In parallel, Wang et al. \cite{wang2024variational} introduced Neural Network Quantum States (NQS) with a ResNet-based amplitude network, demonstrating superior optimization capabilities by achieving improved ground-state energy estimates over complex-valued neural networks in the Heisenberg model. Choe et al. \cite{choe2023can} have proposed  Quantum Genetic Algorithm-based backpropagation (QGA-BP) optimization of neural networks achieves faster convergence and higher correction rates compared to classical backpropagation. 

The application of quantum-enhanced learning mechanisms has continued to evolve. Gupta et al. \cite{gupta2024multiple} proposed the Multiple Controlled Toffoli-driven Adaptive Quantum Neural Network (MCT-AQNN), effectively capturing temporal fluctuations in workload data to improve prediction accuracy. Additionally, Huang et al. \cite{huang2022learning} introduced a meta-learning variational quantum algorithm (meta-VQA), which integrates classical recurrent units to optimize quantum circuit parameters. This approach, leveraging quantum stochastic gradient descent and adaptive learning rates, demonstrated significant efficiency gains in reducing sampling requirements. Seshadri et al. \cite{10508064} further contributed to workload forecasting by presenting a Super Markov Prediction Model (SMPM), which dynamically adapts to changing workload patterns, mitigating resource over- and under-provisioning risks. The comparative performance of classical and quantum neural networks has been explored in \cite{8822629}, revealing that quantum models require fewer training epochs and smaller architectures while achieving comparable or superior results, particularly in computing entanglement witnesses. These findings emphasize the potential of quantum neural networks (QNNs) in surpassing classical approaches for both conventional and quantum-native problems, setting the stage for further advancements in quantum-enhanced predictive modeling. A key comparison of the proposed solution with the state-of-the-art works is summarized in Table \ref{rel-tab:1}.

\begin{table*}[!htbp]
	\caption{Comparison: CA-QNN versus state-of-the-art approaches} \label{rel-tab:1}
	\centering
	\resizebox{0.8\textwidth}{!}{
		\begin{tabular}{p{60pt}|p{80pt}p{80pt}p{66pt}ccc|ccc}
			\hline
			\multirow{3}{*}{\textbf{Models}} & \multicolumn{6}{c|}{\textbf{Key Features}} &\multicolumn{3}{c}{\textbf{Performance Metrics}} \\
			\cline{2-10} & \textbf{Learning} &\textbf{Training} & \multirow{2}{*}{\textbf{Dataset}} &\textbf{Prediction} &\textbf{Adaptive} &\textbf{Structural} & \multirow{2}{*}{\textbf{RMSE}} & \multirow{2}{*}{\textbf{MAE}} & \multirow{2}{*}{\textbf{MAPE}} \\ 
			&\textbf{Approach} &\textbf{Algorithm}& &\textbf{Accuracy} &\textbf{Prediction} & \textbf{Optimization} \\ \hline
			BPNN \cite{lu2016rvlbpnn} & Neural Network &  Back-propagation &  NASA &  $\Downarrow$ & $\times$ & $\times$ &  $\surd$ &  $\times$ & $\times$ \\ \hline
			SaDE \cite{kumar2018workload} &Neural Network & Self adaptive DE & NASA, Saskatchewan & $\downarrow$ &$\surd$ &$\times$ &$\surd$ &$\times$ &$\times$ \\ \hline
			BaDE \cite{kumar2020biphase} &Neural Network & Two phase adaptive DE & NASA, GCD, Saskatchewan & $\Leftrightarrow$ &$\surd$ &$\times$ &$\surd$ &$\times$ &$\times$ \\ \hline
			CLIN \cite{kim2020forecasting} &Traditional regression method & Ensemble & GCD, HPC, Wikipedia web & $\uparrow$ &$\surd$ &$\times$ &$\surd$ &$\times$ &$\surd$ \\ \hline
			LSTM-RNN \cite{kumar2018long} &Neural Network & Adaptive DE  & NASA, Calgary, Saskatchewan & $\uparrow$ & $\surd$ &$\times$ &$\surd$ &$\times$ &$\times$ \\ \hline
			RL-RP \cite{li2021reinforcement} & Reinforcement Learning &Clustering & CPU, LLC &$\uparrow$ &$\times$ &$\times$ &$\times$ &$\times$ &$\times$ \\ \hline
			EQNN \cite{singh2021quantum} &Quantum Neural Network & Self balanced DE & NASA, GCD, HPC, Saskatchewan  & $\uparrow$ &$\surd$ &$\times$ &$\surd$ &$\surd$ &$\times$ \\ \hline
			MCT-AQNN \cite{gupta2024multiple} &MCT Quantum Neural Network & Uniformly adaptive learning & NASA, GCD, Saskatchewan  & $\Uparrow$ &$\surd$ &$\times$ &$\surd$ &$\surd$ &$\surd$ \\ \hline
			meta-VQA \cite{huang2022learning} & Variational Quantum Algorithm (VQA) &Quantum Gradient Descent & ---  &$\Uparrow$ &$\surd$ &$\times$ &$\surd$ &$\times$ &$\times$ \\ \hline
			SMPM \cite{10508064} & Non-Homogeneous Markov Model &Embedding, Score Function & GCD, ACT, TPC-W  &$\Uparrow$ &$\surd$ &$\times$ &$\surd$ &$\surd$ &$\times$ \\ \hline
			QCFNN \cite{kumar2023quantum} &  QNN &  Differential Evolution &  Smart Grid dataset & $\Uparrow$ &  $\surd$ &  $\times$ &  $\surd$ &  $\surd$ &  $\surd$ \\ \hline 
			NAS \cite{zhang2023enhanced} &  Neural networks &  Evolutionary Multi-objective Optimization &  CIFAR100 & $\Uparrow$&  $\surd$ &  $\surd$&  $\times$ &  $\times$ &  $\times$ \\ \hline
			NQS \cite{wang2024variational}  &  Convolutional neural network & Variational optimization &  --- & --- & $\surd$ &  $\times$ &  $\times$ &  $\times$ &  $\times$ \\ \hline
			QGA-BP \cite{choe2023can} &  Neural networks &  Quantum GA-based backpropagation &  ---  & --- & $\surd$ & $\times$ &  $\times$ &  $\times$ &  $\times$\\ \hline 
			
			\textbf{CA-QNN} & Quantum Neural Network & Comprehensively adaptive DE & GCD, ACT & $\Uparrow$ &$\surd$ &$\surd$ &$\surd$ &$\surd$ &$\surd$ \\ \hline
	\end{tabular}}
	\\ \footnotesize{$\Downarrow$: Very low, $\downarrow$: Low, $\Leftrightarrow$: Medium, $\uparrow$: High, $\Uparrow$: Very High, DE: Diﬀerential Evolution, RMSE: Root Mean Square Error, MAE: Mean Absolute Error, MAPE: Mean Absolute Percentage Error, GCD: Google Cluster Data, ACT: Alibaba Cluster Traces, HPC: High Performance Cluster Data; ---: Not Available}
\end{table*}

\subsection{Research Gaps and Motivation for CA-QNN}

Conventional machine learning models, including those in \cite{kumar2018workload, lu2016rvlbpnn, kumar2020biphase, kumar2018long, kim2020forecasting, li2021reinforcement}, rely exclusively on real-valued weight optimization. This limits their ability to capture complex interdependencies in dynamic workloads, leading to low prediction accuracy \cite{wei2021deep, tian2023recent, narayanan2000quantum, kouda2004multilayered, singh2021quantum}. Specifically, real-numbered optimization struggles to model nonlinear feature interactions, hindering adaptability and generalization in high-variance cloud environments. 

Quantum Machine Learning (QML) models overcome these limitations by leveraging quantum properties such as superposition and entanglement to encode richer data representations \cite{wei2021deep, kouda2002image, zhou2007quantum, matsui2000neural}. This makes them particularly well-suited for learning and forecasting dynamic cloud workloads. However, existing QML models face notable challenges. For example, although quantum backpropagation methods offer faster convergence and higher correction rates \cite{choe2023can}, traditional quantum backpropagation lacks efficiency in training deeper networks. Furthermore, as noted in \cite{qian2022dilemma}, current QNNs are constrained by limited model capacity and low sensitivity to regularization, leading to poor generalization and instability across diverse workloads. Our prior works, EQNN \cite{singh2021quantum} and MCT-AQNN \cite{gupta2024multiple}, introduced adaptive evolutionary optimization to improve workload prediction. While effective in leveraging qubit diversity, these models retained fixed architectures, limiting their ability to adapt to changing workload patterns and data heterogeneity.

To address these gaps, we propose the \textbf{C}omprehensively \textbf{A}daptive Architectural Optimization-based \textbf{Q}uantum \textbf{N}eural \textbf{N}etwork (\textbf{CA-QNN}) for dynamic cloud workload prediction and resource allocation. CA-QNN integrates a novel Variable Size-Adaptive Recombination  mechanism, enabling simultaneous optimization of both network parameters and architecture. Coupled with Quantum Adaptive Modulation (QAM), the model dynamically adjusts its topology and learning strategy in real time. This dual adaptability enhances both exploration and precision, significantly improving performance. Unlike existing models, CA-QNN’s concurrent structural and parametric optimization delivers robust and scalable solutions for heterogeneous cloud environments. 
\subsection{Paper contributions}

The major contributions of the proposed work are threefold:

\begin{itemize}
	\item A novel variable-size learning-ingrained QNN named as CA-QNN prediction model is proposed to accurately predict the highly variable and uncertain demands of heterogeneous cloud resources. This model leverages the Controlled-NOT quantum gate for enhanced learning control and utilizes the 360-degree rotational capabilities of qubits to capture the non-linearities and dynamic changes in cloud workloads.
	
	\item We introduced a comprehensively adaptive QNN architecture optimization algorithm for dynamic structural and parametric learning in variable-sized QNNs. This novel approach enhances prediction accuracy and precision by enabling all-inclusive multi-dimensional learning from diverse data samples through optimized architectural organization and inter-connection weights.
	
	\item The implementation with a comprehensive evaluation and comparison with the state-of-the-art approaches confirms the potentiality of the proposed model in terms of its accuracy and efficacy using various workload traces collected from real cloud applications, including Google cluster workloads and Alibaba cluster workloads.  
	
\end{itemize} 

\subsection{Paper Organization} The remaining paper is structured as follows: Section II entails  description of proposed CA-QNN prediction model with its architectural design and qubit vector processing (Section \ref{CA-QNN:Architectural Design}) and structure and parameters optimisation in Section \ref{CA-QNN:Structure and Parameters Optimisation}.  Section \ref{performance evaluation} discusses experimental set-up details and  performance evaluation revealing the simulation and comparison results. The conclusive remarks and future scope are provided in Section \ref{conclusion}. The list of notations with their descriptions used throughout the manuscript is summarized in Nomenclature. 

\section{CA-QNN Model}
\begin{figure*}[!htbp]
	\centering
	\includegraphics[width = 0.9\textwidth]{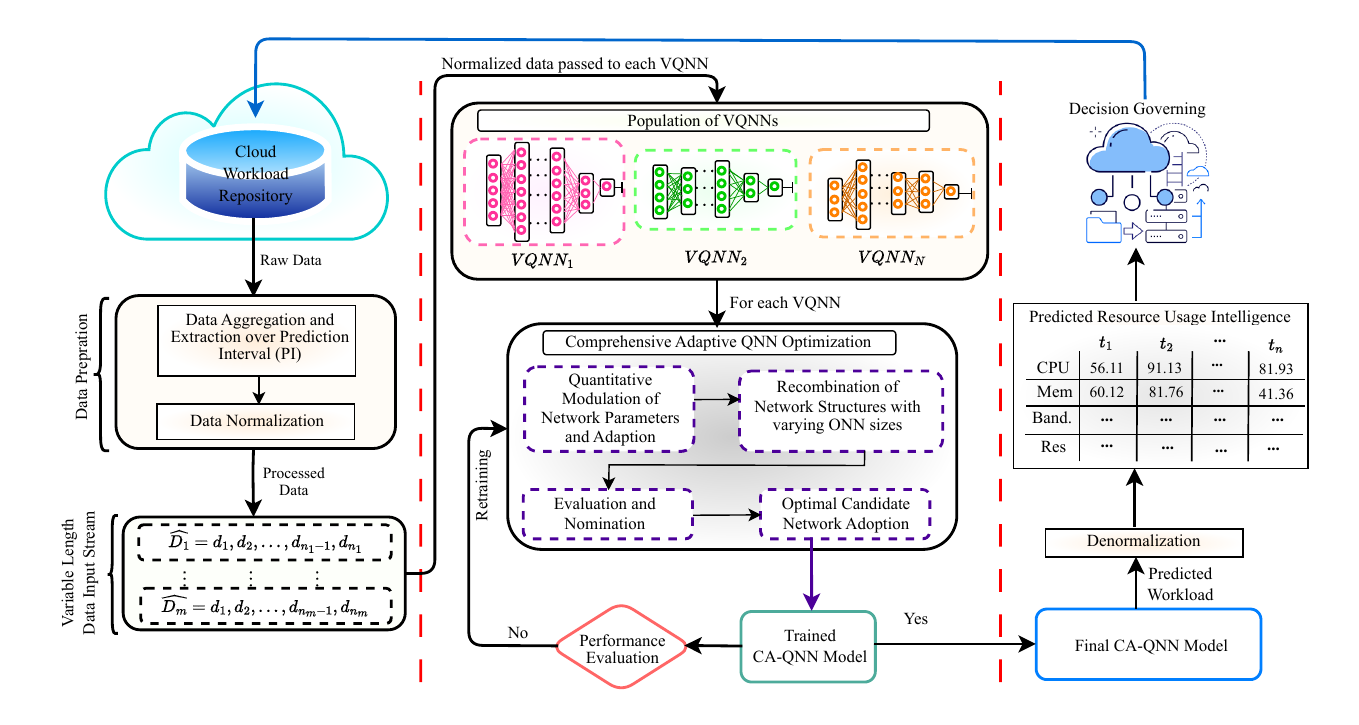}
	\caption{Schematic operational view of CA-QNN model}
	\label{fig1}
\end{figure*}

Consider a cloud data center comprising of $P$ servers: \{$S_1$, $S_2$, ..., $S_P$\} hosting $Q$ virtual machines: \{$V_1$, $V_2$, ..., $V_Q$\} as illustrated in Fig. \ref{fig1}. 
The corresponding  workload allocation and execution information (such as CPU, memory, bandwidth usage, or number of requests over time) is represented as raw data stream of $m^{\ast}$ samples: \{${D}_1$, ${D}_2$, ..., ${D}_{m^{\ast}}$\}, which is captured and maintained at \textit{cloud workload repository}. The raw historical information-driven input data stream is retrieved from the repository. The extraction and time-series-based data aggregation are performed on the raw data  to prepare \(m\) useful data samples, aggregated according to time intervals (e.g., 5 minutes, 10 minutes, 60 minutes, etc.). The \textit{aggregated data samples} are transformed into \textit{normalized data values} using Eq. (\ref{eq-normalisation}), where \(D_i\) and \(\widehat{D_i}\) represent the \(i^{{th}}\) input data and the normalized data values, respectively, and \(D_{{max}}\) and \(D_{{min}}\) are the maximum and minimum data values, respectively. The normalized data $\{\widehat{D}$ =$\widehat{D_1}$, $\widehat{D_2}$, \ldots, $\widehat{D_m}$\} is re-organized  in the form of an
input data matrix ($\widehat{D^{IN}}$) and an output matrix ($\widehat{D^{OP}}$) as shown in Eq. (\ref{eq-input matrix}), where, $n$ + ${x} $ represents $m$, $\widehat{D^{IN}}$ and $\widehat{D^{OP}}$ comprised of $x$ instances of $n$ past values and corresponding actual outcomes, respectively. 

\begin{equation}
	{\widehat{D_i} = \frac{D_i - D_{min}}{D_{max}-D_{min}}}
	\label{eq-normalisation}
\end{equation}
\begin{equation} 
	\resizebox{0.950\hsize}{!}{$\widehat{D^{IN}}= \left[ {\begin{array}{cccc}\widehat{D_1} & \widehat{D_2} & \ldots . & \widehat{D} _n\\ \widehat{D_2} & \widehat{D_3} & \ldots . & \widehat{D_{n+1}} \\ . & . & \ldots . & . \\ . & . & \ldots . & . \\ \widehat{D}_{x} & \widehat{D_{{x}+1}} & \ldots . & \widehat{D_{n+{x}-1}} \\ \end{array} } \right] \widehat{D^{OP}}= \left[ {\begin{array}{c}\widehat{D_{n+1}} \\ \widehat{D{n+2}} \\ . \\ . \\ \widehat{D_{n+{x}}} \end{array} } \right]$}
	\label{eq-input matrix}
\end{equation}

Let  $Pn$ be a population of $N$ randomly generated \textit{Variable Quantum Neural Networks} (VQNNs). Using Eq. (\ref{eq-input matrix}), $N$ \textit{variable-size input vector data streams} are produced by varying the value of $n$ such that $\widehat{D^{1}}$ = \{$\widehat{{D^{1}}_1}$, $\widehat{{D^{1}}_2}$, \ldots, $\widehat{{D^{1}}_{n_1}}$\}; $\widehat{D^{2}}$ = \{$\widehat{{D^{2}}_1}$, $\widehat{{D^{2}}_2}$, \ldots, $\widehat{{D^{2}}_{n_2}}$\}; and $\widehat{D^{N}}$ = \{$\widehat{{D^{N}}_1}$, $\widehat{{D^{N}}_2}$, \ldots, $\widehat{{D^{N}}_{n_N}}$\}, where  \{$\widehat{{D^{1}}}, \widehat{{D^{2}}}, \ldots, \widehat{{D^{N}}}$\} $\in \widehat{D}$.
These data streams \(\{\widehat{D^1}, \widehat{D^2}, \ldots, \widehat{D^N}\}\) are then fed as input data streams to the \(N\) \textit{Variable Quantum Neural Networks} (VQNNs): \{$VQNN_1$, $VQNN_2$, ..., $VQNN_N$\}, respectively for the training and learning process. Specifically, the normalized data \(\widehat{D^1}\) is fed as training input data to  \(VQNN_1\), i.e., \(\widehat{D^1} \mapsto  VQNN_1\); \(\widehat{D^2} \mapsto  VQNN_2\); and \(\widehat{D^N} \mapsto  VQNN_N\).

The detailed description of architecture of VQNN with intended \textit{quantum information processing} is provided in Section \ref{CA-QNN:Architectural Design}. Each VQNN undergoes a systematic structural and network parameters optimization process, referred to as  \textit{Comprehensively Adaptive VQNN Architecture Optimization}, as elaborated in Section \ref{CA-QNN:Structure and Parameters Optimisation}. This learning or optimization process  helps to update the population of VQNNs by executing consecutive steps, including \textit{quantitative modulation of network parameters} (Section \ref{CA-QNN:Adaptive Modulation}), \textit{recombination of network structures with varying configurations} (Section \ref{CA-QNN:Variable Size-Adaptive Recombination}), \textit{fitness evaluation and nomination of candidate VQNNs} (Section \ref{CA-QNN:Evaluation and Nomination}), followed by \textit{optimal VQNN adoption from the population} (Section \ref{CA-QNN:Optimal Candidate Adoption}). Thereafter, a \textit{trained CA-QNN Model} is obtained, which is further evaluated in terms of forecasting error for desired performance verification using fitness evaluation function (Eq. \ref{rmse1}), where, $\digamma(\uplus)$ represents fitness of VQNN, $\widehat{(D)}_i$ and $\widehat{{Fore}(D)_i}$ represent actual and predicted value respectively, and $m$ represents the number of data samples. 

\begin{equation} 
	\digamma(\uplus) = \sqrt{\frac{1}{m}\sum _{i=1}^{m}{(\widehat{(D)}_{i}-\widehat{{Fore}(D)_i}})^2}\label{rmse1} \end{equation}

If the model produces the desired accurate results, a trained CA-QNN model is built, which is further evaluated using unseen testing workload data samples to construct the final CA-QNN model for deployment at the domain expert site. The forecasted data values thus obtained, \(\widehat{{Fore}(D)} = \{\widehat{{Fore}(D)}_{m+1}, \widehat{{Fore}(D)}_{m+2}, \ldots, \widehat{{Fore}(D)}_{m+m^{\ast}}\}\), are denormalized to obtain the forecasted values: \(\mathcal{DEN}(\widehat{{Fore}(D)}) \mapsto {Fore}(D)\), where \({Fore}(D) = \{{Fore}(D)_{m+1}, {Fore}(D)_{m+2}, \ldots, {Fore}(D)_{m+m^{\ast}}\}\). This model forecasts workload status, including expected resource demands, number of requests, and future load capacity, and guides effective resource management decisions within the data center. It aims to minimize resource wastage, electrical power consumption, and carbon emissions, while suggesting optimized load scheduling and management to make effective utilization of available resources.

\subsection{ Architectural Design and Qubit Vector Processing} \label{CA-QNN:Architectural Design} 
\begin{figure*}[!htbp]
	\centering
	\includegraphics[width = 0.85\linewidth]{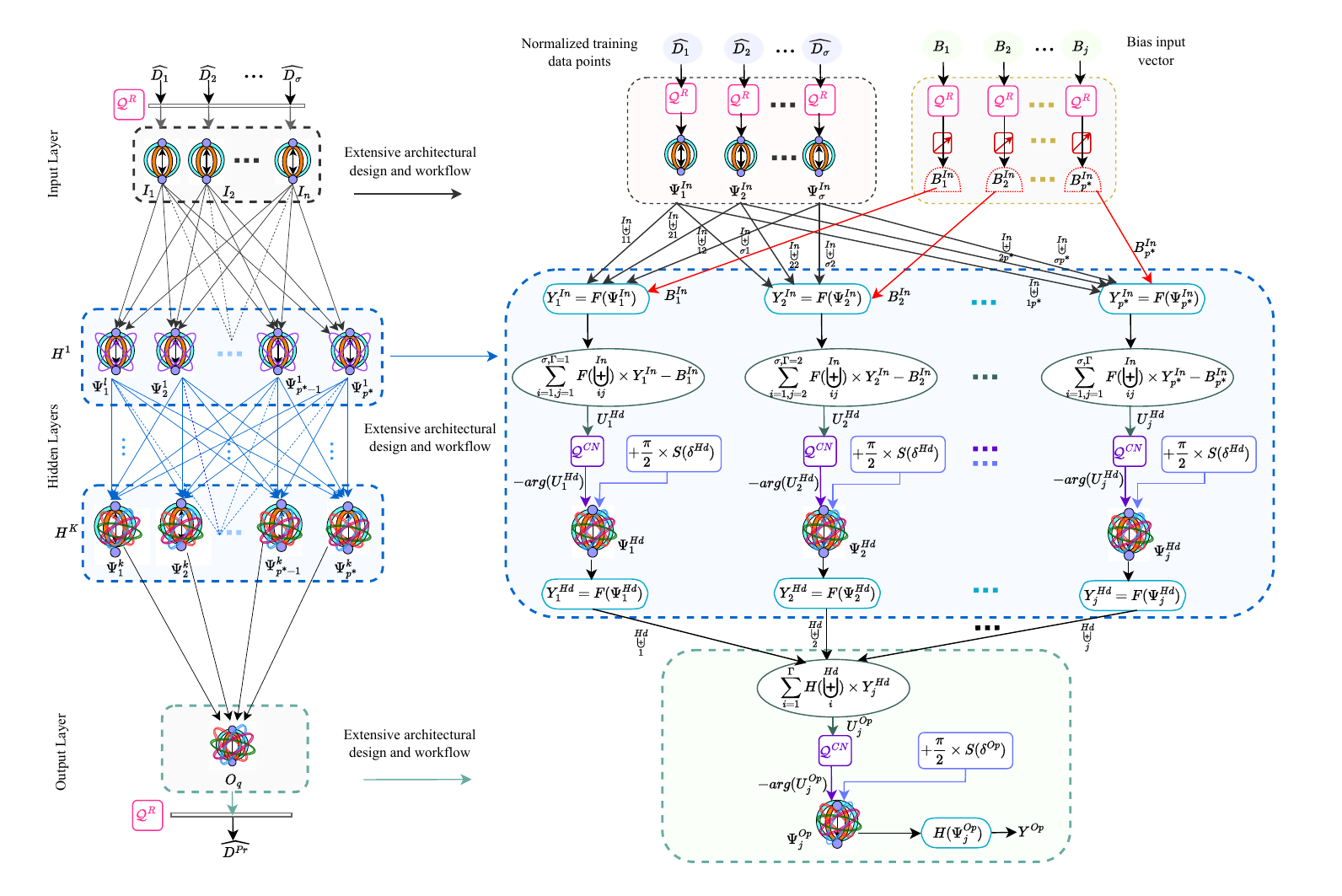}
	\caption{CA-QNN architectural design and qubit vector processing}
	\label{QNN2}
\end{figure*}

The abstract and concrete architectural design of the CA-QNN model is illustrated in Fig. \ref{QNN2}. It is a multi-layered, variable-sized quantum neural network (VQNN) consisting of three key layers: an input layer with \( n \) qubit neurons \( I = \{I_1, I_2, \ldots, I_n\} \), multiple hidden layers with varying \( p^\ast \) qubit neurons \( H^l = \{H^l_1, H^l_2, \ldots, H^l_{p^\ast}\} \) for \( l \in \{1, 2, ..., K\} \), and an output layer containing a single qubit neuron \( O_q \), where \( q = 1 \).  Each qubit neuron in a given layer is connected to neurons in the subsequent layer through variable quantum neural weights \( \uplus^{(\ell)i}_{(\ell+1)j} \), which dynamically adjust during the comprehensive architectural learning process (Section \ref{CA-QNN:Structure and Parameters Optimisation}).
Unlike classical networks, where weight adjustments occur in discrete steps, CA-QNN leverages quantum superposition and phase transformation, allowing it to encode and process multiple workload states simultaneously. This enhances adaptability by enabling the model to efficiently handle diverse and rapidly fluctuating cloud workloads without requiring explicit retraining for each variation. Additionally, the quantum state’s ability to represent complex probability distributions improves the model’s responsiveness to fluctuations in workload intensity, resource demand, and latency constraints. By integrating phase encoding and amplitude modulation, CA-QNN offers a flexible and scalable approach to managing nonlinear workload variations.

As qubits propagate from the \( \ell^{th} \) layer to the \( (\ell+1)^{th} \) layer, they undergo transformations governed by the \textit{quantum rotation gate} (\( \mathcal{Q}^R \)) and the \textit{quantum C-Not gate} (\( \mathcal{Q}^{CN} \)). The \( \mathcal{Q}^R \) gate facilitates controlled phase adjustments, optimizing qubit interactions for enhanced learning precision. Meanwhile, the \( \mathcal{Q}^{CN} \) gate regulates qubit flow, ensuring adaptive learning across dynamically changing workloads. By processing information in a qubit vector form and leveraging these quantum gates, CA-QNN efficiently captures intricate workload patterns, offering a robust and computationally efficient solution for cloud resource management. The normalized training data points: $\widehat{D}$ = \{$\widehat{D_1}$, $\widehat{D_2}$, \ldots, $\widehat{D_n}$\} are then transformed into a qubit input vector, denoted as $\{{\Psi_1^{In}}, {\Psi_2^{In}}, \ldots, \Psi_n^{In}\} \in \Psi^{In}$, applying the effect of the $\mathcal{Q}^R$ gate using Eqs. (\ref{eq5}) and (\ref{eq6}) in the input layer.  The term $F(.)$ represents a quantum activation function or a nonlinear transformation function applied to the input quantum state ($\Psi^{In}_{i}$). The \textit{bias input vector}  $B^{In} = \{B^{In}_1, B^{In}_2, \ldots, B^{In}_p\}$ is produced using Eq. (\ref{eq:bias}), where  $R_j$ is a random value such that $R_j\in [-1, 1] \subset \mathbb{R}$. 

\begin{gather}
	{Y}_{i}^{In}= {{F}}(\Psi^{In}_{i}) \label{eq5} 
	\\
	\Psi^{In}_{i}=\frac {\pi }{2}\times \widehat{D_i} \label{eq6}    
	\\
	B^{In}_j = \frac{\pi}{2} \times  R_j  \quad \forall j \in \{1, 2, \ldots, p\} \label{eq:bias}
\end{gather}

The input qubit vector \(\{\Psi_1^{{In}}, \Psi_2^{{In}}, \ldots, \Psi_n^{{In}}\}\) is processed through the hidden layers using the \(\mathcal{Q}^{{CN}}\) gate. This processing generates the intermediate qubit vector  \(Y_j^{{Hd}}\) as described by Eqs. \eqref{eq7}, \eqref{eq8}, and \eqref{eq9}. The adjusted qubit vector \(\Psi_j^{{Hd}}\) is then derived by applying the reverse rotation or non-rotation effect of the C-NOT gate, which is controlled by the reversal parameter \(\varrho\), to the magnitude of the intermediate qubit vector  \(\arg(U_j)\). 
\begin{gather} 
	U_{j}^{Hd}=\sum _{i=1,j=1}^{n}F(\uplus_{ij}^{In})\times  Y_i^{In}-B^{In} \label{eq7} \\  \Psi_{j}^{Hd}= \frac{\pi }{2}\times {\mathcal{SG}}(\varrho^{Hd}) - arg( U_{j}^{Hd}) \label{eq8}\\ Y_{j}^{Hd}= F(\Psi_j^{Hd})\label{eq9} 
	\end{gather}
	The magnitude  of $j^{th}$ qubit vector  ($\arg(U_{j}^{Hd})$) in  \( z \)-dimensional Hilbert space is given by Eq. \eqref{a1}; where, \( u_i \) represents the complex probability amplitude of the \( i^{th} \) basis state, and its modulus is computed using Eq. \eqref{a2}:
	
	\begin{gather} \label{a1}
\|U^{Hd}\| = \sqrt{\sum_{i=1}^{z} |u_i|^2}
\\
\label{a2}
|u_i| = \sqrt{\text{Re}(u_i)^2 + \text{Im}(u_i)^2}
\end{gather}
where, $\text{Re}(u_i)$  and $\text{Im}(u_i)$ specify the Real  and Imaginary values, respectively, associated with the qubit (represents as a complex number). For a properly normalized qubit vector, the magnitude satisfies \( \|U^{Hd}\| = 1 \). The sigmoid function  \(\mathcal{SG}(\varrho)\), which outputs a value between 0 and 1, is defined in Eq. \eqref{eq10}.
\begin{gather} 
\mathcal{SG}(\varrho) = \frac{1}{1 + exp^{(-\varrho)}}\label{eq10} 
\end{gather}

Similarly, the qubit vector  \(Y_j^{{Op}}\) in the output layer is generated using Eqs. \eqref{eq11}, \eqref{eq12}, and \eqref{eq13}.

\begin{gather}  U_{j}^{Op}=\sum _{i=1}^{p} H(\uplus_{i}^{Hd})\times  Y_{j}^{Hd} \label{eq11} \\
\Psi_{j}^{Op}= \frac{\pi }{2}\times \mathcal{SG}(\varrho^{Op}) - arg( U_{j}^{Op}) \label{eq12}\\
Y_{}^{Op}=  H(\Psi_j^{Op})\label{eq13} \end{gather}

\subsection{Structure and Parameters Optimization} \label{CA-QNN:Structure and Parameters Optimisation}
The comprehensive structure and inter-connection parameters optimization of CA-QNN model is accomplished in four consecutive steps including (\textit{i}) \textit{Quantum Adaptive Modulation}; (\textit{ii})  \textit{Variable Size-Adaptive Recombination} \cite{Lotika-4630614}; (\textit{iii}) \textit{Evaluation and Nomination};  and (\textit{iv}) \textit{Optimal Candidate Adoption} over population of  $N$ VQNN candidate vectors. Let \{$\uplus_1$, $\uplus_2$, ..., $\uplus_N$\} constitute the randomly generated population of VQNN candidate vectors. The performance of each VQNN vector is evaluated using fitness evaluation function (Eq. \ref{rmse1}) to locate the initial optimal candidate vector ($\uplus_{Opm}$). 

\subsubsection{Quantum Adaptive Modulation} \label{CA-QNN:Adaptive Modulation} The population of $N$ VQNN vectors undergoes adaptive exploration and exploitation by randomly selecting one among three quantum adaptive modulation strategies to produce $N$ \textit{adaptively modulated VQNN vectors} \{$\Delta_1$, $\Delta_2$, ..., $\Delta_N$\}. The
three intended quantum adaptive modulation strategies are   \textit{Quantum Adaptive/Randomized Modulation} ($QARM$),  \textit{Quantum Adaptive/Current-to-Optimal Modulation} ($QACO$), and   \textit{Quantum Adaptive/Optimal Modulation} ($QAOM$), which are stated in Eq.\eqref{eq16}, Eq.\eqref{eq17}, and Eq.\eqref{eq18}, respectively,  where $R^{mod}$ represents rate of modulation.
\begin{gather}  \Delta_{i}^{j}=\uplus_{r_{1}}^{j} +  R^{mod}_{i} \times \left ({\uplus_{r_{2}}^{j} -\uplus_{r_{3}}^{j}}\right ) \label{eq16}\\
\begin{split}
	\Delta_{i}^{j}=\uplus_{i}^{j} +  R^{mod}_{i} \times \left ({\uplus_{Opm}^{j} -\uplus_{i}^{j}}\right ) \\
	+  R^{mod}_{i} \times \left ({\uplus_{r_{1}}^{j} -\uplus_{r_{2}}^{j}}\right ) \qquad \label{eq17}
\end{split}
\\
\Delta_{i}^{j}=\uplus_{Opm}^{j} +  R^{mod}_{i} \times \left ({\uplus_{r_{1}}^{j} -\uplus_{r_{2}}^{j}}\right )  \label{eq18}\end{gather}

The terms: $\Delta_{i}^{j}$ and $\uplus_{i}^{j}$ are the $i^{th}$ modulated and current qubit vectors of $j^{th}$ iteration from VQNN population space $Pn$, respectively, and $\uplus_{Opm}^{j}$ is the optimal solution found so far, and mutually distinct three random numbers ($r_1, r_2$ and $r_3 $) are generated in the range [1, ${N}$]. The probabilities for selecting $QARM$, $QACO$, and $QAOM$ modulation techniques are denoted by  $T_{1}$, $T_{2}$, and  $T_{3}$, respectively. Initially, these probabilities are set to  $T_{1} = T_{2} = 0.33$ and  $T_{3} = 0.34$, ensuring that each technique has an approximately equal chance of being selected. The \textit{roulette wheel selection} method is then used to choose among these three modulation strategies based on the assigned probabilities. To implement this selection process,  a \textit{modulation strategy selection} ($mss$) vector  consisting of random numbers (between 0 and 1) equals to size of the intended VQNN, denoted as $ L = |\uplus|$, is produced.  The selected modulation strategy is denoted as $ M^{\ast}$ in Eq. (\ref{eq19}). This approach allows for dynamic adjustment of the selection probabilities for the different modulation techniques based on their previous performance, thereby enhancing the overall exploration process.

\begin{gather} 
M^{\ast } =
\begin{cases} QARM, & {\text {If}\left ({0 < mss_{i} \leq T_{1} }\right )} \\ QACO, & {\text {If}\left ({T_{1} < mss_{i} \leq T_{1} + T_{2} }\right )} \\ QAOM,  & {\text {Otherwise}} \end{cases} \label{eq19}
\end{gather}

To evaluate the effectiveness of each modulation strategy, the variables $\varepsilon_{1}$, $\varepsilon_{2}$, and $\varepsilon_{3}$ to represent the number of successful candidates selected for the next generation through the $QARM$, $QACO$, and $QAOM$ modulation strategies, respectively. Similarly, let $\tau_{1}$, $\tau_{2}$, and $\tau_{3}$ denote the number of candidates that failed to progress to the next generation for each respective strategy. These variables allow to measure and compare the performance of each strategy and adopt the most admissible modulation strategy. The probabilities:  $T_{1}$, $T_{2}$, and  $T_{3}$, which influence the exploration and exploitation processes for the $QARM$, $QACO$, and $QAOM$ modulation techniques, respectively, are computed using Eqs. \eqref{eq:T1}-\eqref{eq:T3}; 

\begin{gather} 
T_{1}=\frac {\varepsilon_{1}\left ({\varepsilon_{2} + \tau_{2} + \varepsilon_{3} + \tau_{3} }\right )}{St} \label{eq:T1}\\ 
T_{2}=\frac {\varepsilon_{2}\left ({\varepsilon_{1} + \tau_{1} + \varepsilon_{3} + \tau_{3} }\right )}{St} \label{eq:T2}\\ 
T_{3}=1- \left ({T_{1} +T_{2} }\right )  \label{eq:T3} 
\end{gather}
where, $St$ is computed as follows:
\begin{align*} 
St =&2(\varepsilon_{2}\varepsilon_{3} + \varepsilon_{1}\varepsilon_{3} + \varepsilon_{1}\varepsilon_{2}) + \tau_{1}(\varepsilon_{2} + \varepsilon_{3}) \\&+ \tau_{2}(\varepsilon_{1} + \varepsilon_{3}) + \tau_{3}(\varepsilon_{1} + \varepsilon_{2})  \label{eqS}
\end{align*}       

\subsubsection{Variable Size-Adaptive Recombination} \label{CA-QNN:Variable Size-Adaptive Recombination}
\begin{figure} [!htbp]
\centering
\includegraphics[width = 0.5\textwidth]{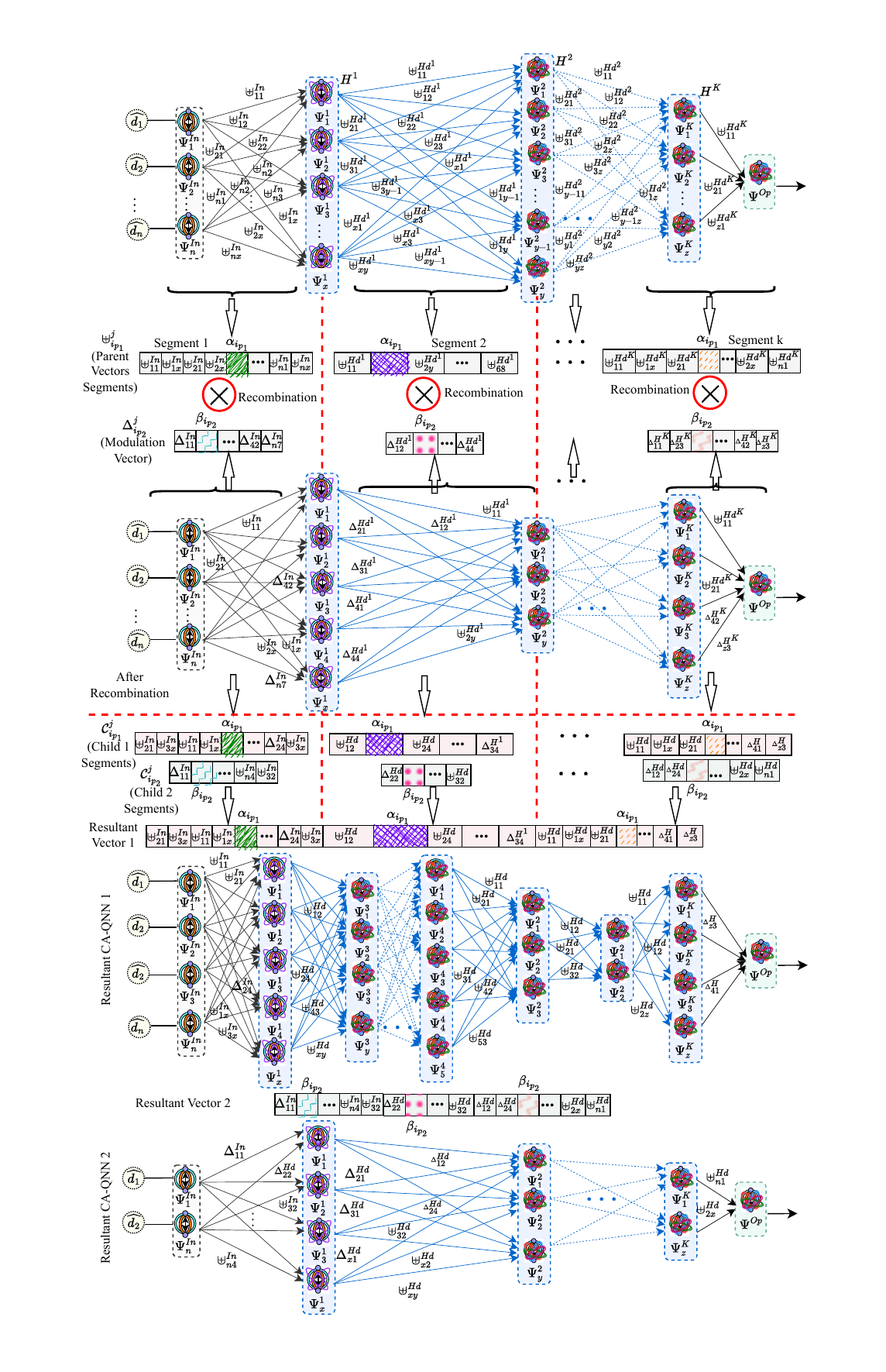}
\caption{Variable size-adaptive recombination}
\label{fig:crossover}
\end{figure}

Within VQNN, the weight connection qubit vector is divided into $K$ segments: I-H connections of length ($n \times p^\ast$);  H-H connections of length ($p^\ast\times p^\ast$); and H-O connections of length ($p^\ast \times q$). Here, $n$ and $q$ are fixed, while $p^\ast$ is variable, causing the segment lengths to vary with $p^\ast$. Fig. \ref{fig:crossover} illustrates the variable-length crossover mechanism.
The $i^{th}$ parent vectors of the $j^{th}$ generation, including \(\uplus^j_{i_{p1}}\) and modulated vector \(\Delta^j_{i_{p2}}\), undergo recombination to produce offspring vectors: \(C^j_{i_{p1}}\) and \(C^j_{i_{p2}}\). Specifically, \(C^j_{i_{p1}}\) is produced by swapping a block segment of \(\uplus^j_{i_{p1}}\) with a corresponding block segment in \(\Delta^j_{i_{p2}}\) marked by the point of recombination (\(\alpha_{i_{p1}} \in [0, L]\)) in \(p1\) using Eq. \eqref{c1}. The second offspring vector, \(C^j_{i_{p2}}\), is developed by exchanging a block segment of \(\Delta^j_{i_{p2}}\) marked at recombination point (\(\beta_{i_{p2}}\)) in \(p2\) with a corresponding block segment from \(\uplus^j_{i_{p1}}\) as specified in Eq. \eqref{c2}.

\begin{equation}
\label{c1}
C^j_{i_{p1}} = \uplus^j_{i_{p1}} || \alpha_{i_{p1}} (\uplus^j_{i_{p1}}\Leftrightarrow \Delta^j_{i_{p2}})
\end{equation}
\begin{equation}
\label{c2}
C^j_{i_{p2}} = \Delta^j_{i_{p2}} || \beta_{i_{p2}} (\uplus^j_{i_{p1}}\Leftrightarrow \Delta^j_{i_{p2}})
\end{equation}
\subsubsection{Evaluation and Nomination} \label{CA-QNN:Evaluation and Nomination}
The fitness of each candidate vector of updated population is evaluated using Eq. (\ref{rmse1}). Both parent and child vectors are nominated and the most admissible and optimal candidate vector is adopted to proceed further with CA-QNN optimization.

\subsubsection{Optimal Candidate Adoption} \label{CA-QNN:Optimal Candidate Adoption}
The population for the next generation (\(j + 1\)) is determined based on Eq. \eqref{eqq25} by adopting suitable candidates (\(\uplus^{j+1}_i : i \in \{1, 2, \ldots, N\}\)) by comparing the fitness values of the current existing solution (\(\uplus^j_i\)) and the offspring vector (\(C^j_i\)).

\begin{equation}
\uplus^{j+1}_i =
\begin{cases} 
	C^j_i & \text{if } \digamma(C^j_i) \leq \digamma(\uplus^j_i) \\
	\uplus^j_i & \text{otherwise}
\end{cases}
\label{eqq25}
\end{equation}

\subsubsection{ Formal Convergence Proof for Quantum Adaptive Modulation (QAM) Optimization}

Let \( \uplus^j_{i} \) be the parameterized qubit vector at iteration \( j \), and let the QAM weight update follow one of three mutation strategies stated as Eq.\eqref{eq16}, Eq.\eqref{eq17}, and Eq.\eqref{eq18}.  The modulation strategy is selected based on probability using Eq. (\ref{eq19}),  where the selection probabilities are dynamically updated using Eqs. \eqref{eq:T1}-\eqref{eq:T3}.  \\
We prove that QAM training is guaranteed to converge under the following conditions:
\begin{itemize}
\item  \textit{Bounded Step Size}: The modulation factor \(  R_i^{mod} \) is constrained within the range \( (0,1) \), ensuring controlled and stable updates.

\item  \textit{Monotonic Fitness Improvement}: The fitness function \( \digamma(\uplus) \) is a non-increasing sequence as stated in Eq. \eqref{f1}:
\begin{equation}  \label{f1}
	\digamma(\uplus_{t+1}) \leq \digamma(\uplus_t)
\end{equation}
\item  \textit{Probability Selection Stabilization}: The probabilities converge to a steady state using Eqs. \eqref{l1} and \eqref{l2}:
\begin{align} \label{l1}
	\lim_{t \to \infty} T_1 = T_1^*, \quad \lim_{t \to \infty} T_2 = T_2^*, \quad \lim_{t \to \infty} T_3 = T_3^*, \\ T_1^* + T_2^* + T_3^* = 1. \label{l2}
\end{align}

As the number of iterations \( t \) increases indefinitely (\( t \to \infty \)), the probabilities \( T_1, T_2, T_3 \) approach fixed values \( T_1^*, T_2^*, T_3^* \), which represent the final stable probabilities of selecting each modulation strategy. Additionally, the condition (Eq. \ref{l2}) ensures that the three probabilities remain valid (i.e., they always sum to 1), as required for a valid probability distribution.

\end{itemize}
The convergence is established using a \textit{Lyapunov stability} argument, showing that the fitness function is a decreasing sequence with a lower bound. The \textit{Lyapunov Function} (Fitness Score)  is defined using Eq. \eqref{rmse}:
\begin{equation} 
Lf(\uplus) = \digamma(\uplus) =\sqrt{\frac{1}{m}\sum _{i=1}^{m}{(\widehat{(D)}_{i}-\widehat{{Fore}(D)_i}})^2}\label{rmse} \end{equation}
For convergence, the constraint is represented using Eq. \eqref{proof_2}, must be specified. 
\begin{equation} \label{proof_2}
Lf(\uplus_{t+1}) - Lf(\uplus_t) < 0, \quad \forall t.
\end{equation}

The expectation ($\mathbb{E}$) over the QAM weight update is stated using Eq. \eqref{proof_3}:
\begin{equation}  \label{proof_3}
\mathbb{E} \left[ Lf(\uplus_{t+1}) \right] = \mathbb{E} \left[ Lf(\uplus_t) - R_i^{mod} \|\uplus^j_{Opm} - \uplus^j_{i} \|^2 \right].
\end{equation}
Since \( 0 < {R_i^{mod}} < 1 \), this ensures \textit{monotonic reduction in fitness} and $T_1$, $T_2$, $T_3$ update using reinforcement, we establish:
\begin{itemize}
\item  \( T_1, T_2, T_3 \) are bounded in \([0,1]\).
\item  The strategy probabilities adjust dynamically to enhance fitness improvement.
\item  Over time, the most effective strategy prevails, as described by Eq. \eqref{proof_4}.
\begin{equation} \label{proof_4}
	\lim_{t \to \infty} M^* = M^*_{\text{optimal}}.
\end{equation}
\end{itemize}

Since \( Lf(\uplus) \) is bounded below (error cannot be negative) and step sizes decrease, Eq. \eqref{proof_5}; which implies that \( Lf(\uplus) \) converges to a finite value, ensuring training stability.
\begin{equation} \label{proof_5}
\sum_{t=0}^{\infty} \left( Lf(\uplus_{t+1}) - Lf(\uplus_t) \right) < \infty.
\end{equation}

\subsection{Algorithm and Complexity}
The operational summary of training mechanism of CA-QNN is conferred in Algorithm \ref{algo1}. 
\begin{algorithm}[!htbp]
\caption{CA-QNN Learning Algorithm}\label{algo1}
Initialize $ R^{mod}_{\mu}$ = 0.5, $ R^{mod}_{n}$=0.3, $T_{1}=T_{2}= 0.33$, $T_{3}$= 0.34 \; 
Generate random population of $N$ VQNN vectors  \;
Evaluate fitness of each VQNN using Eq. \eqref{rmse1} \;
\For{ each generation ${t\leq \mathcal{G}}$ }{
	Generate $mss_i$ vector for adoption of adaptive modulation \;
	\For{ each VQNN $i$ }{
		Generate $r_1 \neq r_2 \neq r_3 \neq i \in [1,L] $ and $j_{rand} \in [1,L]$ \;
		\If {0 $<$ $mss_i \leq T_{1}$ }{
			Apply $QARM$ \;
		}
		\ElseIf {$T_{1}<mss_i \leq T_{1} + T_{2}$}{
			Apply $QACO$ } 
		\Else {
			Apply $QAOM$} 
	}
	Divide the modulated vector in $K$ segments to perform variable-sized adaptive recombination \;
	
	\textit{For first segment of size: $n\times p$}, where $p$ is number of nodes in first hidden layer: Recombination point is located at $ n \times ((int)(rand) \times \lvert p_{1} - p_{2} \rvert)$
	to produce child segments, where, $p_{1}$ \& $p_{2}$ are number of nodes in first hidden layer of parent 1 and parent 2 \;
	\textit{For second segment of size: $p \times p^\ast$}:
	Recombination will follow: $ p \times ((int)(rand) \times \lvert p^\ast_{1} - p^\ast_{2} \rvert)$ to produce other child segment \;
	
	Combine relevant segments and evaluate fitness using Eq. \eqref{rmse1}  \;
	Select or adopt the most admissible candidate and proceed for next generation \;
}
\end{algorithm} 
The optimization process comprises consecutive steps including generation of random population of VQNNs, fitness evaluation for $m$ input data samples, and intermediate vector generation by quantum adaptive modulation followed by variable size-adaptive recombination with time complexities with $\mathcal{O}(1)$, $\mathcal{O}(SL) = \mathcal{O}(n^2L)$ (since $S = p^\ast(n + 1)$ and assuming $p^\ast \equiv n$, implying $S \equiv n^2$), $\mathcal{O}(m)$, and $\mathcal{O}(L)$, respectively. This process is repeated for $t$ generations, leading to an overall time complexity of $\mathcal{O}(n^2mtL)$. The input data, population, and intermediate vectors are stored as an $m \times n$ matrix, an $L \times S \equiv L \times n^2$ matrix, and $\mathcal{O}(L)$ vectors, respectively, resulting in a total space complexity of $\mathcal{O}(n^2L + mn)$.

\section{Performance Evaluation} \label{performance evaluation}
\subsection{Experimental Set-up}
The simulation experiments are conducted on a server machine assembled with two Intel\textsuperscript{\textregistered} Xeon\textsuperscript{\textregistered} Silver 4114 CPU with a 40 core processor and 2.20 GHz clock speed. The computation machine is deployed with 64-bit Ubuntu 16.04 LTS, having 128 GB RAM.  The proposed work is implemented in Python 3.7 with the details listed in Table \ref{table:name1}.

\begin{table}[ht]
\caption{ Experimental set-up parameters with values}
\centering
\resizebox{1.0\columnwidth}{!}{
	\begin{tabular}{lc|lc}
		\hline
		Parameter & Value &Parameter & Value\\
		\hline
		Input neural nodes ($n$)    & 10 & Iterations  & 50\\
		Hidden layer nodes ($p^{\ast}$)    & 5-10 &Total epochs      & 10-50 \\
		Output layer nodes ($q$)   & 1 & Architectures & 4\\
		Size of Training set   & 60\%-80\% & Best Candidates & 4 \\
		Size of Test set  & 40\%-20\% &Activation & Sigmoid\\
		Population size   & 80 & Number of hidden layers& 1-4\\
		\hline
\end{tabular}}
\label{table:name1}
\end{table}

\subsection{Data Sets}
The performance analysis and comparison of CA-QNN model is conducted using 
four different benchmark datasets including \textit{Cluster workloads}: CPU and memory usage traces from  Google Cluster Data (GCD) \cite{Reiss2011} and \textit{Alibaba  Cluster  traces} (ACT) obtained from \cite{9068614}. 
GCD workload provides behavior of cloud applications for cluster and big data analytics such as Hadoop which gives resource usage traces collected over a period of 29 days. ACT comprises  4,000 machines, captured runtime resource usage over an 8-day period. From this dataset, a random selection of 1,000 machines over the same 8 days is made, with each machine containing approximately 20,000 traces. The CPU and memory utilization of VM traces  are extracted from both of these datasets and are aggregated over the period of first ten days for different prediction intervals (PI) like, 5 minutes, 10 minutes, ..., 60 minutes (1 hour), 1440 minutes (1 day). Table \ref{tab:ds} shows the statistical characteristics of the evaluated workloads. 
\begin{table}[!htbp]
\caption{Characteristics of evaluated workloads}\label{tab:ds}
\centering
\resizebox{1.0\columnwidth}{!}{
	\begin{tabular}{lccccccc}
		\hline
		\multirow{2}{*}{Workload} & \multirow{2}{*}{Duration} & \multirow{2}{*}{Origin} &No. of & Machines & Traces/ & \multirow{2}{*}{Mean} & \multirow{2}{*}{Std.Dev.} \\ 
		&&& Machines & Selected & Machine && \\
		\hline
		GCD-CPU & 29 days& 2011 & 125K & 1000 & 100K & 21.84& 13.62 \\ \hline
		GCD-Mem & 29 days& 2011 & 125K & 1000 & 100K & 19.55& 16.6 \\ \hline
		ACT-CPU & 8 days& 2019 & 4K & 1000 & 20K & 28.575 & 20.499 \\ \hline
		ACT-Mem & 8 days& 2019 & 4K & 1000 & 20K & 91.114 & 4.577 \\ 
		\hline
\end{tabular}}
\end{table}

\subsection{Results}
The extensive evaluation of the CA-QNN model across various ratios of training and testing data sizes (60:40, 70:30, and 80:20) is summarized in Table \ref{result-tab:1}, using GCD-CPU workload data traces. The experiments were conducted for each error metric individually, and their corresponding training times are also reported.
\begin{table*}[!htbp]
\caption{ Performance evaluation of CA-QNN for prediction over different training testing scenarios} \label{result-tab:1}
\centering
\resizebox{0.95\textwidth}{!}{
	\begin{tabular}{c|c|l|lll|lll|lll}
		\hline
		\multirow{2}{*}{\textbf{DT}} & \textbf{Error} & \textbf{PI} & \multicolumn{3}{c|}{\textbf{Training (60\%) / Testing (40\%)}} &\multicolumn{3}{c|}{\textbf{Training (70\%) / Testing (30\%)}} &\multicolumn{3}{c}{\textbf{Training (80\%) / Testing (20\%)}} \\
		\cline{4-12} & \textbf{Metrics}&\textbf{(minutes)}&\textbf{Training} & {\textbf{Testing}} & {\textbf{Time (ms)}}  & \textbf{Training} & {\textbf{Testing}} & {\textbf{Time (ms)}} & \textbf{Training} &{\textbf{Testing}} & {\textbf{Time (ms)}} \\ \hline
		\multirow{18}{*}{\rotatebox{90}{GCD-CPU}}& \multirow{6}{*}{\rotatebox{90}{RMSE}} &5	&0.07822	&0.04245& 314645&	0.07551&	0.08233 & 366113 &	0.07983&	0.07249 & 416770 \\ 
		&	& 10	& 0.04651	&0.04244&159236	&0.04714	&0.03831 &182823	&0.04737	&0.03389 &	210093 \\
		&	&20	&0.05303&	0.05444 &80301	&0.05380	&0.04887 &	92679 &	0.05388	& 0.05250 &106287 \\
		&	&30	&0.11610	&0.11321 &53536 &	0.11463	 &0.11492 &60966	&0.11068&	0.11866 &71162 \\
		&	&60	&0.11616	&0.10021 &26683 &	0.12775	&0.11737 & 30602 &	0.11935	&0.13497 &35184 \\
		&	&1440	&0.09964 &	0.10194 & 1155&	0.10359	&0.08128 &1332	&0.08561	&0.11676 &1458 \\ 
		\cline{2-12}	& \multirow{6}{*}{\rotatebox{90}{MAE}} & 5	&0.04479	&0.04495 &314414	&0.04839	&0.04786 &364454	&0.04552	&0.04839 &416544 \\
		&&	 10&	0.01619	&0.016112 &158791	&0.02267	&0.02210 &183922	&0.02292	&0.02289 &210291 \\
		&& 20 &	0.02483	&0.02680 &79626	&0.02541	&0.02625 &	92423	&0.02558	&0.02666 &105549 \\
		&& 30&	0.07414	&0.07479 &53059	&0.07549	&0.07231 &60585	&0.07740	&0.07245 &70573 \\
		&&60	&0.07914	&0.07239 &26753	&0.08145	&0.07542 &31424	&0.07989	&0.08146 &35411 \\
		&&1440	&0.07801	&0.12033 &1138	&0.07535	&0.10700 &1349	&0.07222	&0.13500 &1419 \\ 
		\cline{2-12}& \multirow{6}{*}{\rotatebox{90}{MAPE}} & 5	&0.53605	&0.53827 &313393	&0.52801	&0.53309 &364533	&0.53757	&0.53783 &	416287 \\
		&&10	&0.67325	&0.65298 &158634	&0.66540	&0.63624 &183062	&0.66447	&0.67748 &209782 \\
		&&20	&0.66929	&0.64745 &	79607	&0.65888	&0.67427 &92296	&0.66037	&0.64291 &104656 \\
		&&30	&0.39723	&0.39499 &52696	&0.07723	&0.07725 &61368	&0.38288	&0.41599 &70031 \\
		&&60	&0.45537	&0.41963 &	26399	&0.41372	&0.44619 &30358	&0.43215	&0.43789 &34653 \\
		&&1440	&0.41585	&0.41372 &1125	&0.08523	&0.11329 &1234	&0.32114	&0.35690 &1403 \\ \hline
\end{tabular}}
\\ \footnotesize{\scriptsize PI: Prediction Interval, DT: Workload Data Traces, GCD-CPU: Google Cluster CPU Traces, ms: mili-seconds}
\end{table*}
The major error metrics considered include \textit{root mean squared error} (RMSE), \textit{mean absolute error} (MAE), and \textit{mean absolute percentage error} (MAPE) over different prediction intervals (PI) ranging from 5 minutes to 1440 minutes. Both training and testing error metrics (RMSE, MAE, MAPE) show an increasing trend as the PI increases from 5 minutes to 1440 minutes across all training and testing size ratios. This indicates that higher PI results in fewer training data samples, leading to higher error rates. Larger training data sizes (e.g., 80:20 ratio) result in slightly lower error metrics compared to smaller training data sizes (e.g., 60:40 ratio). This improvement in accuracy comes at the cost of increased training time. Specifically, the 80:20 ratio consumes approximately 10.3\% more training time than the 60:40 ratio.
The CA-QNN model demonstrates consistent performance across different training and testing data size ratios. The model achieves higher accuracy with larger datasets, although this requires more training time. The CA-QNN model performs well across a wide range of PI, from short-term (5 minutes) to long-term (1 day) predictions, regardless of the training data sizes. This versatility confirms the robustness and adaptability of CA-QNN model to various prediction scales. Therefore,  CA-QNN model delivers promising accuracy across different PI and training data sizes, with a slight trade-off in training time for larger datasets. This makes the CA-QNN model a reliable choice for various prediction scenarios.
\par 
Table \ref{res-tab:2} presents the prediction accuracy of the CA-QNN model for GCD-Mem, ACT-CPU, and ACT-Mem workload data traces across a range of prediction intervals (PI) from 5 minutes to 1440 minutes, using a training to testing data size ratio of 60:40.  The results show that the values of RMSE, MAE, and MAPE are higher for ACT workloads due to the smaller number of data samples compared to GCD workload traces. The majority of the error metrics follow the reduction trend in the order: ACT-CPU $\equiv$ ACT-Mem $\leq$ GCD-Mem $\equiv$ GCD-CPU (analysed from Table  \ref{result-tab:1}). Overall, RMSE, MAE, and MAPE fall within the ranges of [0.01-0.12], [0.01-0.14], and [0.1-0.7], respectively. These results demonstrate the robust and appealing performance of the CA-QNN model, regardless of the training data samples and PI. 



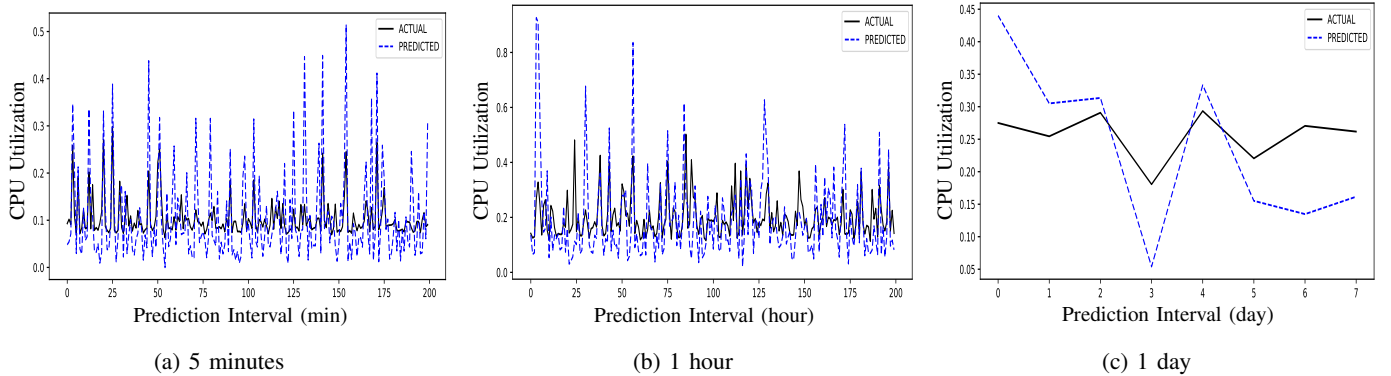
\begin{figure*}[!htbp]
\centering
\begin{subfigure}[t]{0.32\textwidth}
	\centering
	\begin{tikzpicture}
		\node[inner sep=0pt] (A) {\includegraphics[width=2.2in,height=4cm]{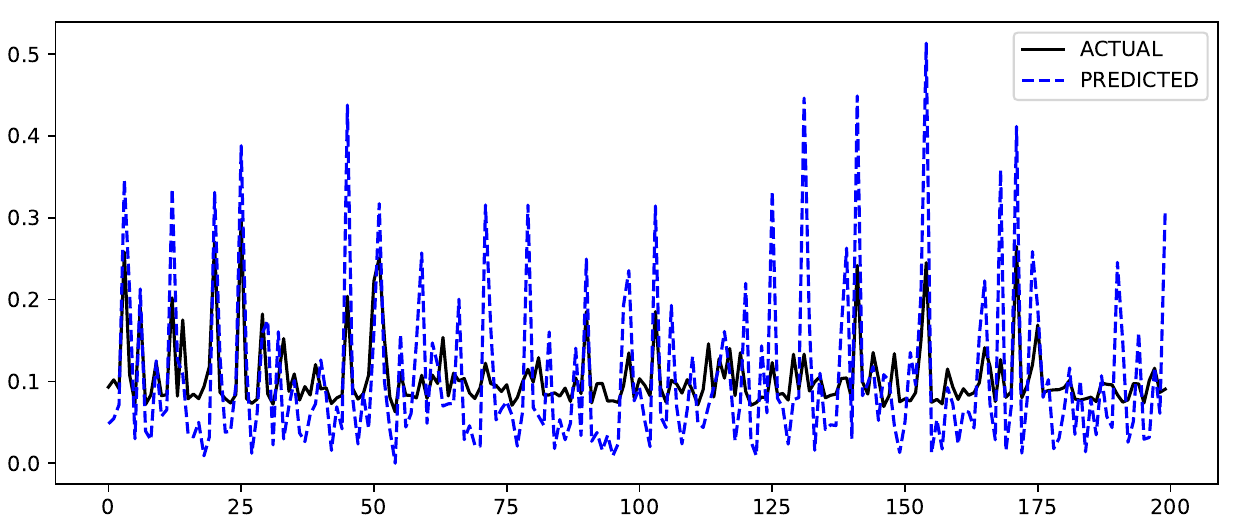}};
		\node[black] (B) at ($(A.south)!-0.05!(A.north)$) {\footnotesize  Prediction Interval (min)};
		\node[black,rotate=90] (C) at ($(A.west)!-0.03!(A.east)$) {\footnotesize  CPU Utilization};
	\end{tikzpicture}
	\caption{5 minutes}
\end{subfigure}
\hfill
\begin{subfigure}[t]{0.32\textwidth}
	\centering
	\begin{tikzpicture}
		\node[inner sep=0pt] (A) {\includegraphics[width=2.2in,height=4cm]{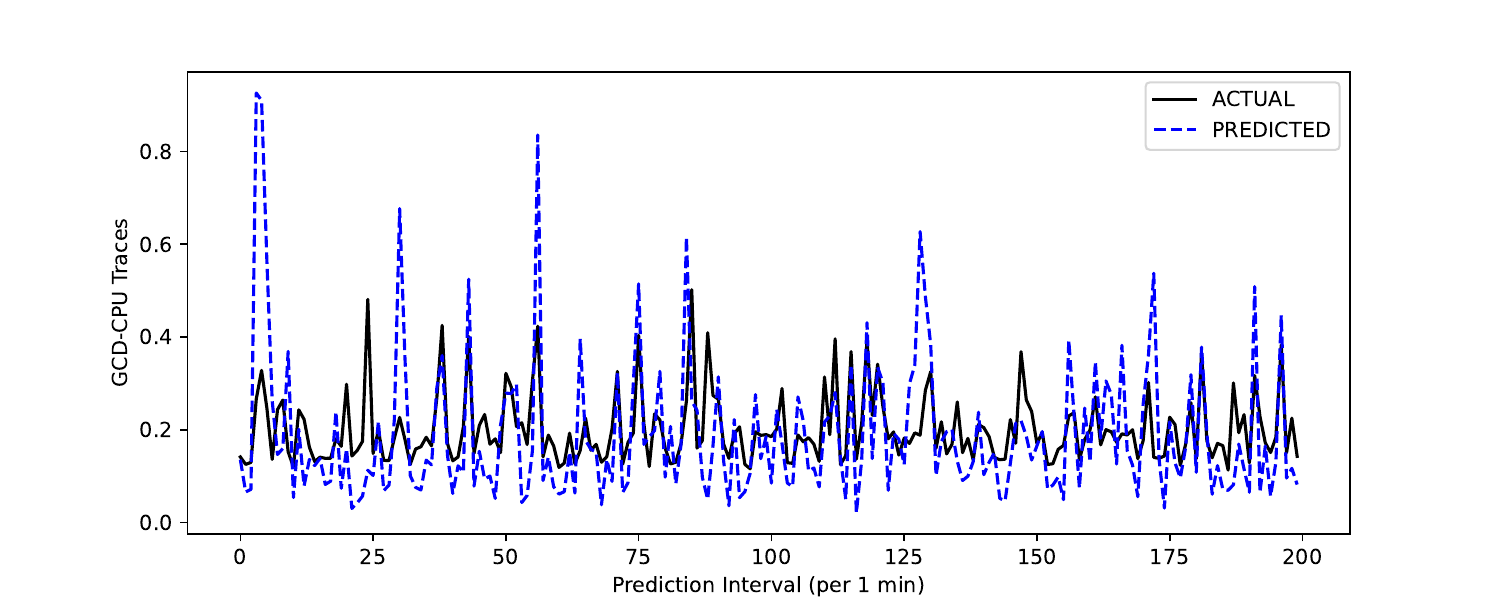}};
		\node[black] (B) at ($(A.south)!-0.05!(A.north)$) {\footnotesize  Prediction Interval (hour)};
		\node[black,rotate=90] (C) at ($(A.west)!-0.03!(A.east)$) {\footnotesize  CPU Utilization};
	\end{tikzpicture}
	\caption{1 hour}
\end{subfigure}
\hfill
\begin{subfigure}[t]{0.32\textwidth}
	\centering
	\begin{tikzpicture}
		\node[inner sep=0pt] (A) {\includegraphics[width=2.2in,height=4cm]{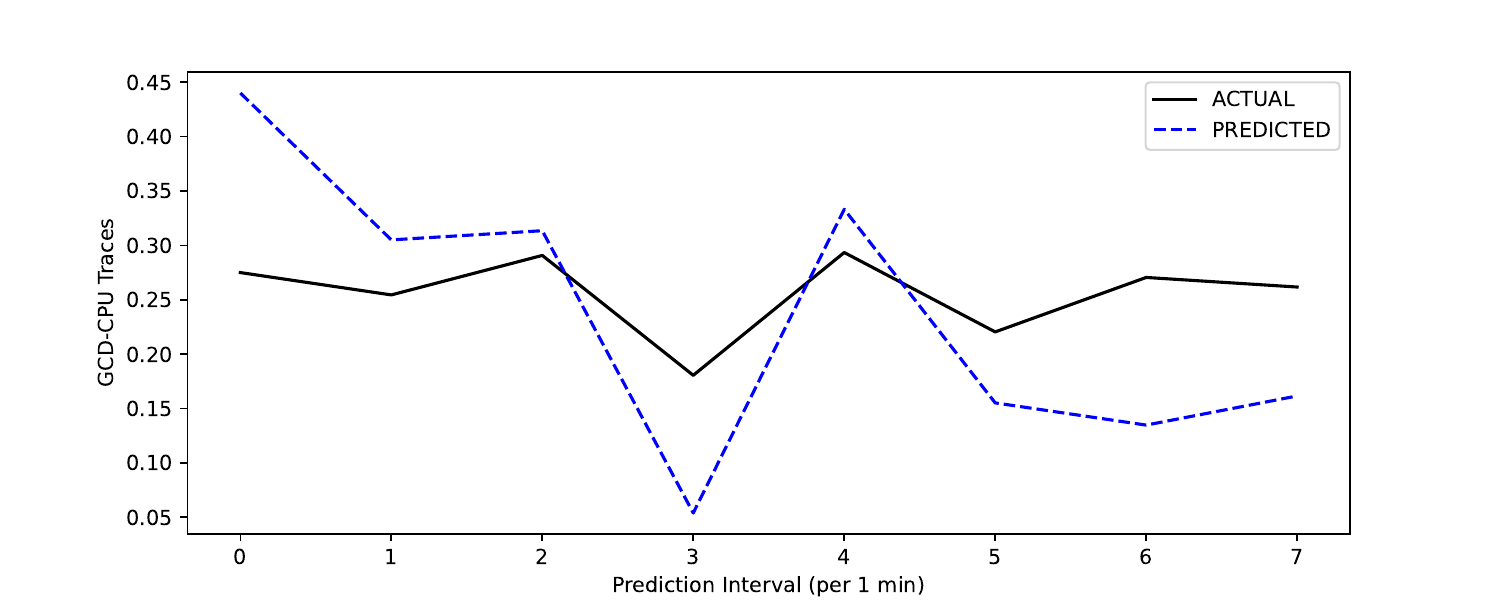}};
		\node[black] (B) at ($(A.south)!-0.05!(A.north)$) {\footnotesize  Prediction Interval (day)};
		\node[black,rotate=90] (C) at ($(A.west)!-0.03!(A.east)$) {\footnotesize  CPU Utilization};
	\end{tikzpicture}
	\caption{1 day}
\end{subfigure}
\caption{Actual versus predicted workload for GCD-CPU}
\label{res-fig:1a}
\end{figure*}

\begin{figure*}[!htbp]
\centering
\begin{subfigure}[t]{0.32\textwidth}
	\centering
	\begin{tikzpicture}
		\node[inner sep=0pt] (A) {\includegraphics[width=2.2in,height=4cm]{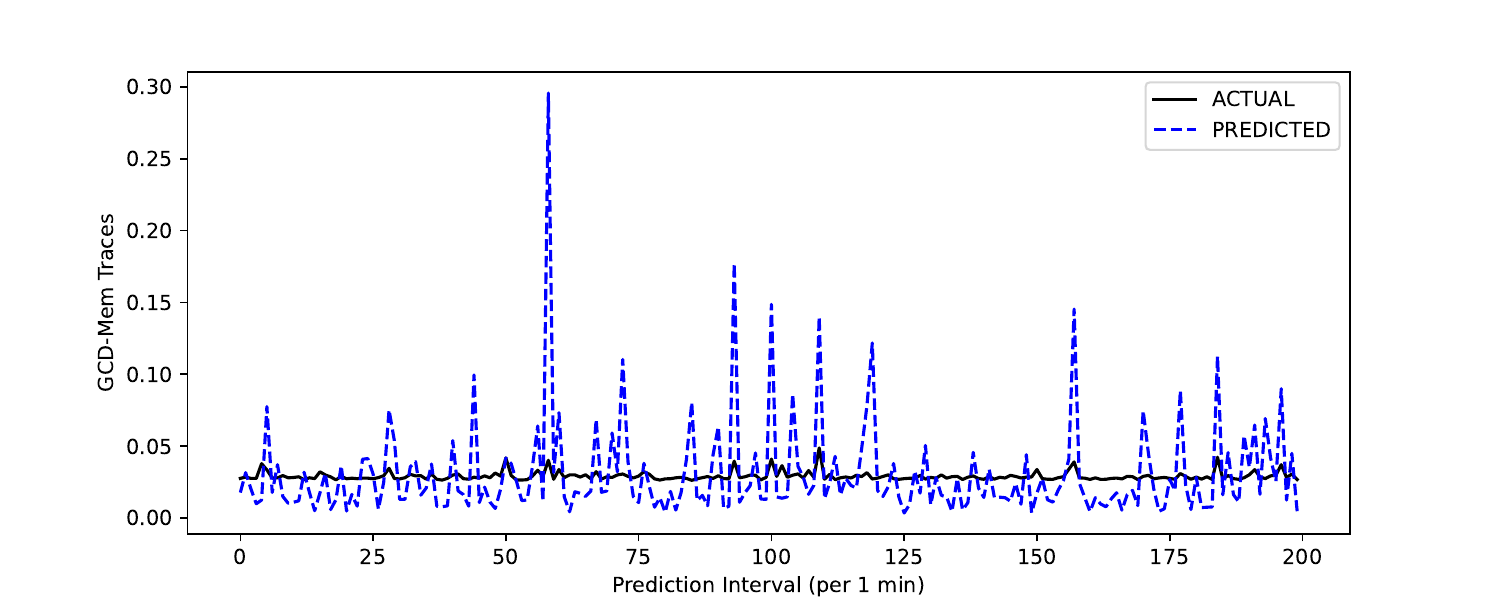}};
		\node[black] (B) at ($(A.south)!-0.05!(A.north)$) {\footnotesize  Prediction Interval (min)};
		\node[black,rotate=90] (C) at ($(A.west)!-0.03!(A.east)$) {\footnotesize  Memory Utilization};
	\end{tikzpicture}
	\caption{5 minutes}
\end{subfigure}
\hfill
\begin{subfigure}[t]{0.32\textwidth}
	\centering
	\begin{tikzpicture}
		\node[inner sep=0pt] (A) {\includegraphics[width=2.2in,height=4cm]{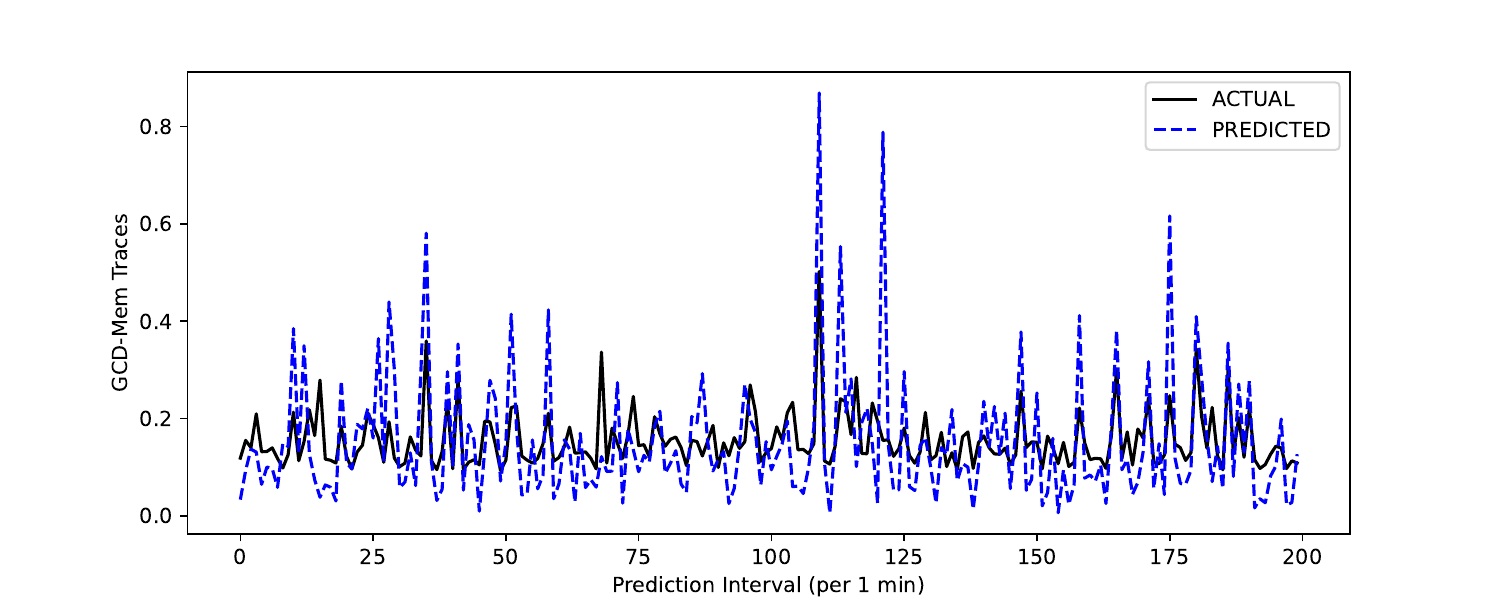}};
		\node[black] (B) at ($(A.south)!-0.05!(A.north)$) {\footnotesize  Prediction Interval (hour)};
		\node[black,rotate=90] (C) at ($(A.west)!-0.03!(A.east)$) {\footnotesize  Memory Utilization};
	\end{tikzpicture}
	\caption{1 hour}
\end{subfigure}	
\hfill
\begin{subfigure}[t]{0.32\textwidth}
	\centering
	\begin{tikzpicture}
		\node[inner sep=0pt] (A) {\includegraphics[width=2.2in,height=4cm]{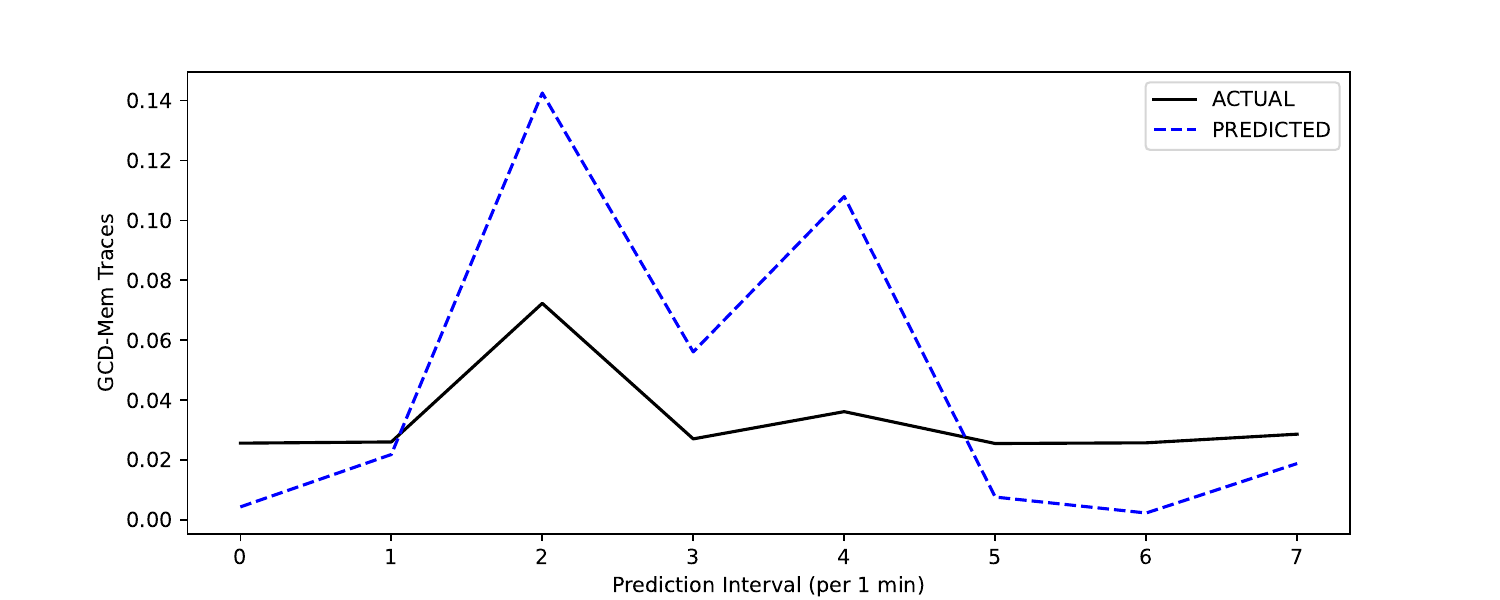}};
		\node[black] (B) at ($(A.south)!-0.05!(A.north)$) {\footnotesize  Prediction Interval (day)};
		\node[black,rotate=90] (C) at ($(A.west)!-0.03!(A.east)$) {\footnotesize  Memory Utilization};
	\end{tikzpicture}
	\caption{1 day}
\end{subfigure}
\caption{Actual versus predicted workload for GCD-Mem}
\label{res-fig:1b}
\end{figure*}

\begin{figure*}[!htbp]
\centering
\begin{subfigure}[t]{0.32\textwidth}
	\centering
	\begin{tikzpicture}
		\node[inner sep=0pt] (A) {\includegraphics[width=2.2in,height=4cm]{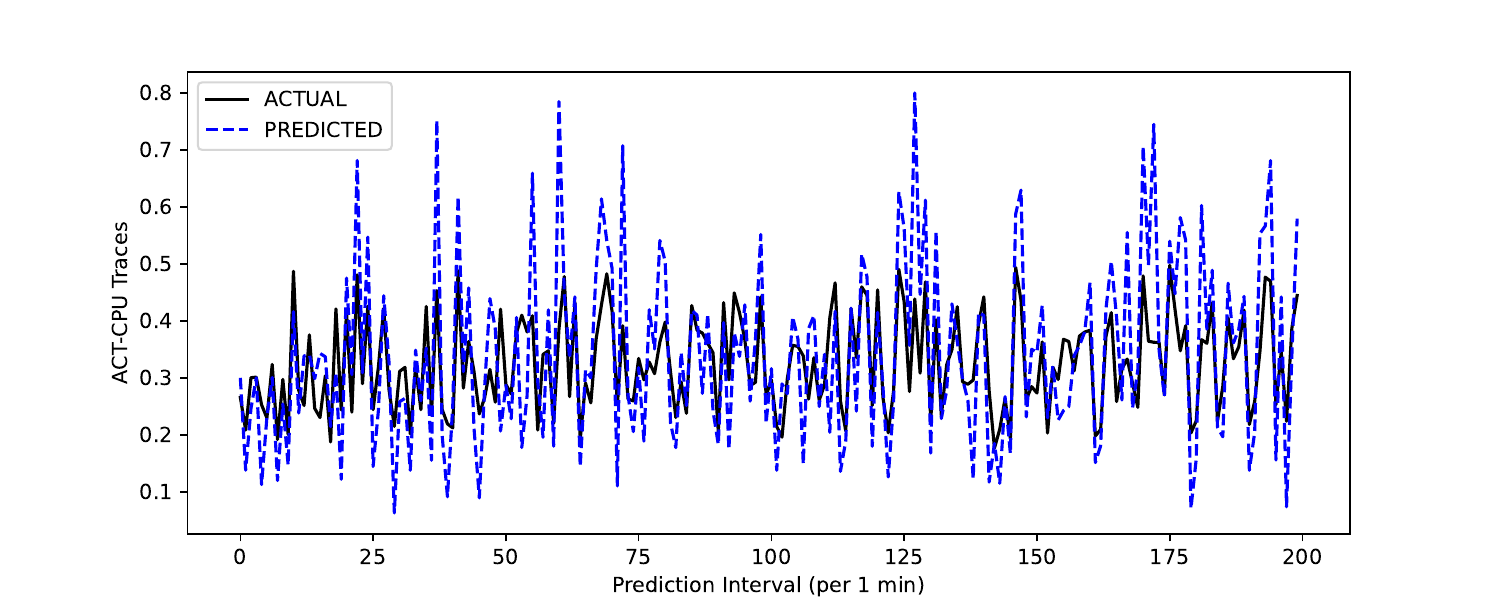}};
		\node[black] (B) at ($(A.south)!-0.05!(A.north)$) {\footnotesize  Prediction Interval (min)};
		\node[black,rotate=90] (C) at ($(A.west)!-0.03!(A.east)$) {\footnotesize  CPU Utilization};
	\end{tikzpicture}
	\caption{5 minutes}
\end{subfigure}
\hfill
\begin{subfigure}[t]{0.32\textwidth}
	\centering
	\begin{tikzpicture}
		\node[inner sep=0pt] (A) {\includegraphics[width=2.2in,height=4cm]{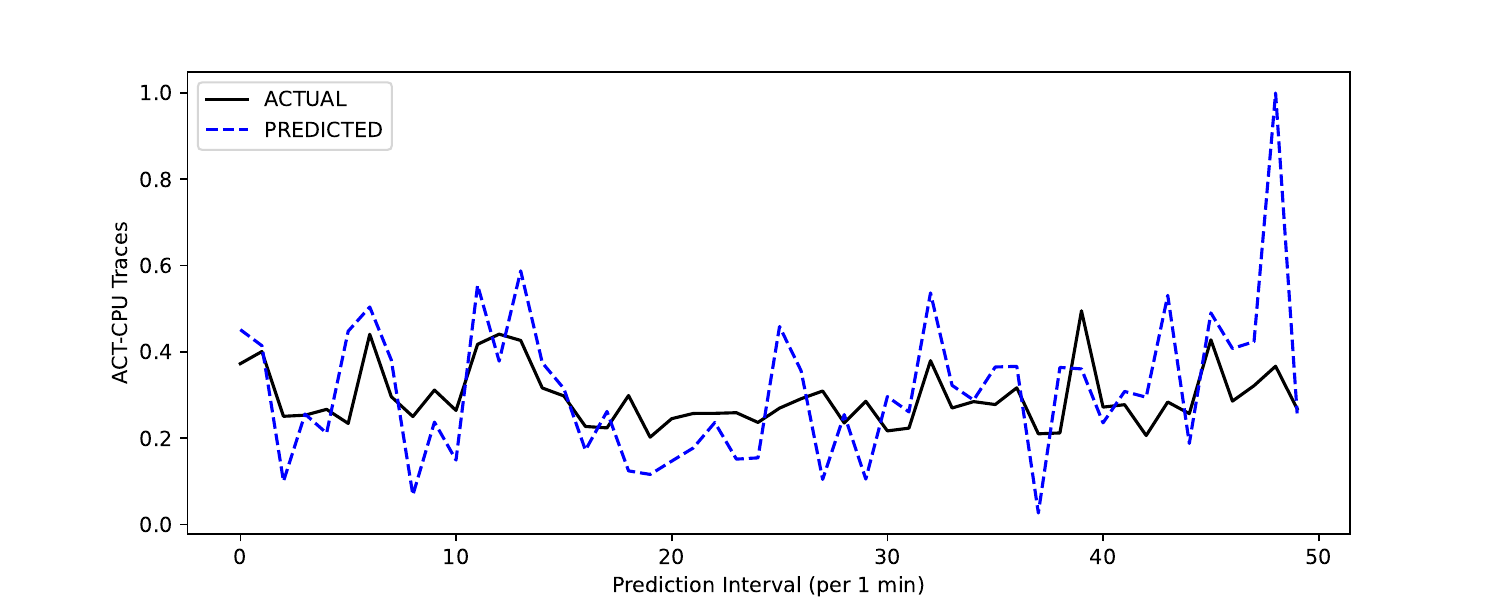}};
		\node[black] (B) at ($(A.south)!-0.05!(A.north)$) {\footnotesize  Prediction Interval (hour)};
		\node[black,rotate=90] (C) at ($(A.west)!-0.03!(A.east)$) {\footnotesize  CPU Utilization};
	\end{tikzpicture}
	\caption{1 hour}
\end{subfigure}
\hfill
\begin{subfigure}[t]{0.32\textwidth}
	\centering
	\begin{tikzpicture}
		\node[inner sep=0pt] (A) {\includegraphics[width=2.2in,height=4cm]{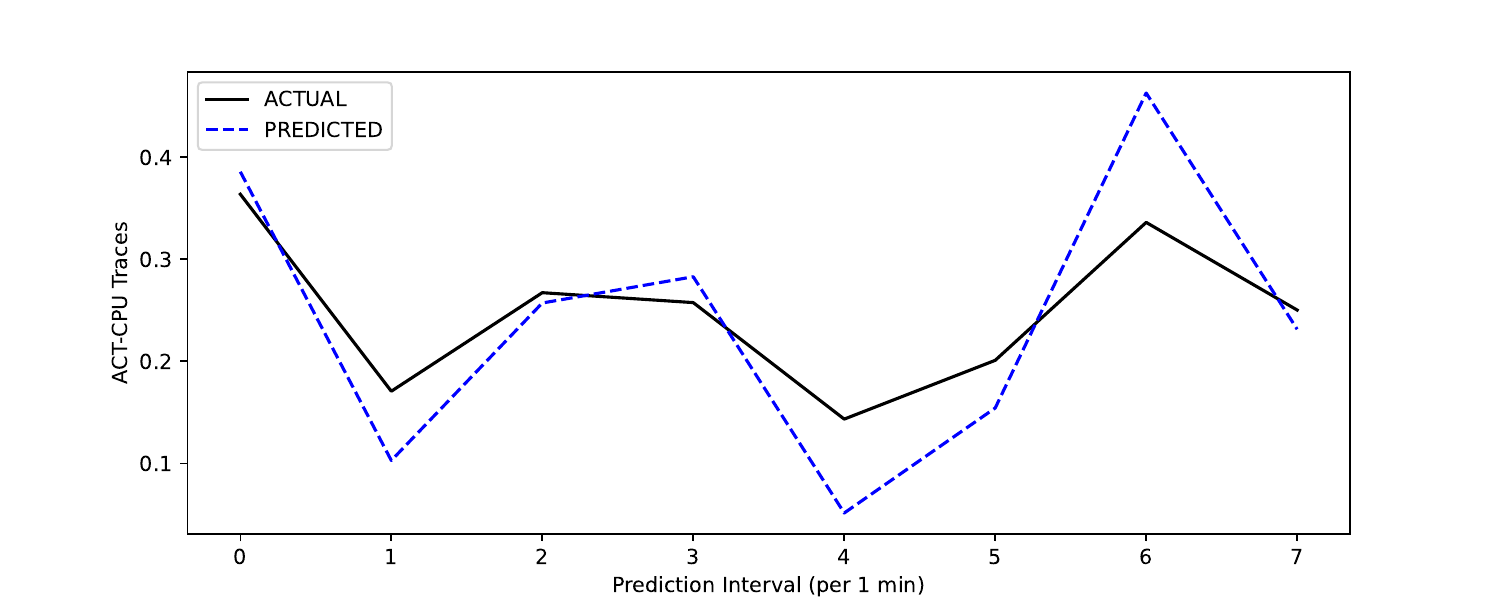}};
		\node[black] (B) at ($(A.south)!-0.05!(A.north)$) {\footnotesize  Prediction Interval (day)};
		\node[black,rotate=90] (C) at ($(A.west)!-0.03!(A.east)$) {\footnotesize  CPU Utilization};
	\end{tikzpicture}
	\caption{1 day}
\end{subfigure}
\caption{Actual versus predicted workload for ACT-CPU}
\label{res-fig:1c}
\end{figure*}

\begin{figure*}[!htbp]
\centering
\begin{subfigure}[t]{0.32\textwidth}
	\centering
	\begin{tikzpicture}
		\node[inner sep=0pt] (A) {\includegraphics[width=2.2in,height=4cm]{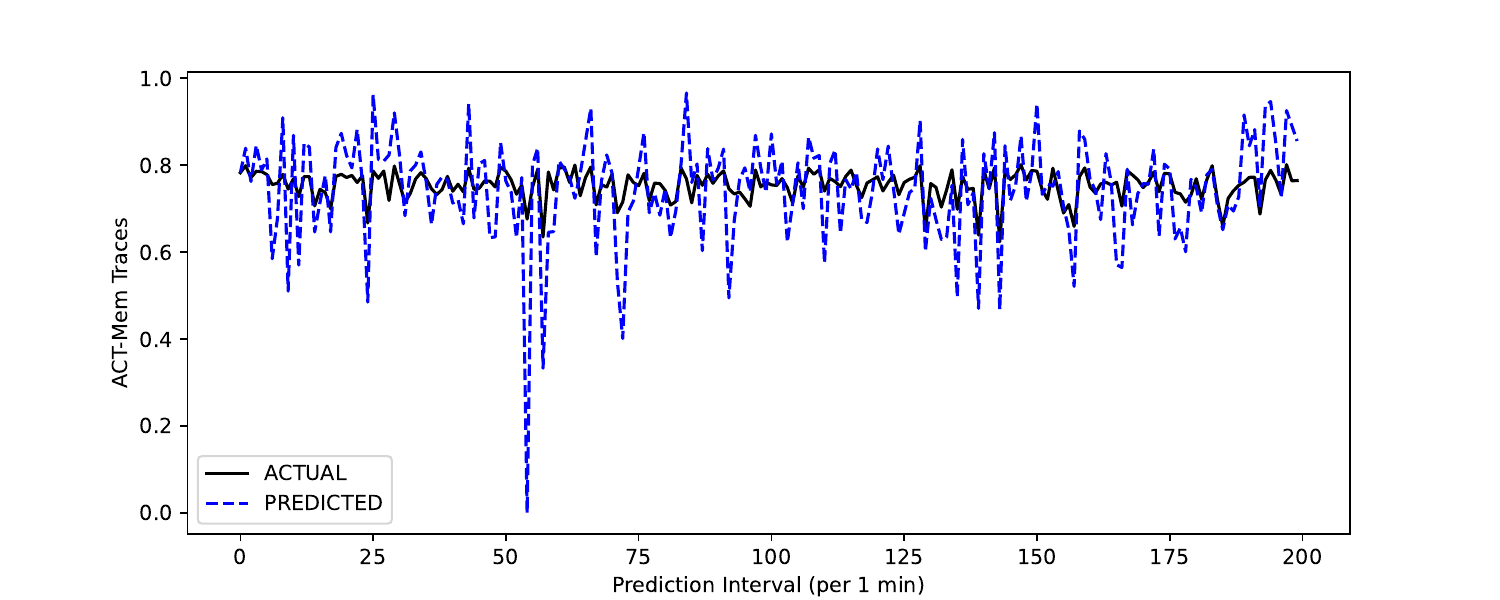}};
		\node[black] (B) at ($(A.south)!-0.05!(A.north)$) {\footnotesize  Prediction Interval (min)};
		\node[black,rotate=90] (C) at ($(A.west)!-0.03!(A.east)$) {\footnotesize  Memory Utilization};
	\end{tikzpicture}
	\caption{5 minutes}
\end{subfigure}
\hfill
\begin{subfigure}[t]{0.32\textwidth}
	\centering
	\begin{tikzpicture}
		\node[inner sep=0pt] (A) {\includegraphics[width=2.2in,height=4cm]{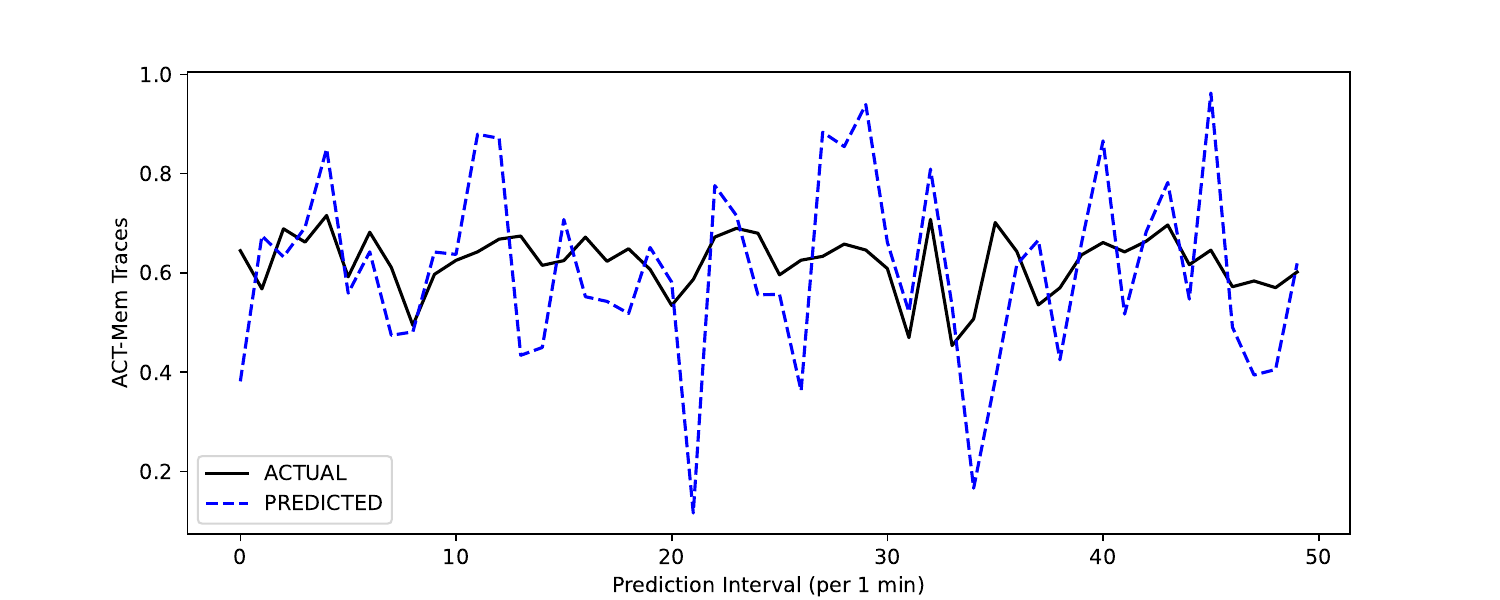}};
		\node[black] (B) at ($(A.south)!-0.05!(A.north)$) {\footnotesize  Prediction Interval (hour)};
		\node[black,rotate=90] (C) at ($(A.west)!-0.03!(A.east)$) {\footnotesize  Memory Utilization};
	\end{tikzpicture}
	\caption{1 hour}
\end{subfigure}	
\hfill
\begin{subfigure}[t]{0.32\textwidth}
	\centering
	\begin{tikzpicture}
		\node[inner sep=0pt] (A) {\includegraphics[width=2.2in,height=4cm]{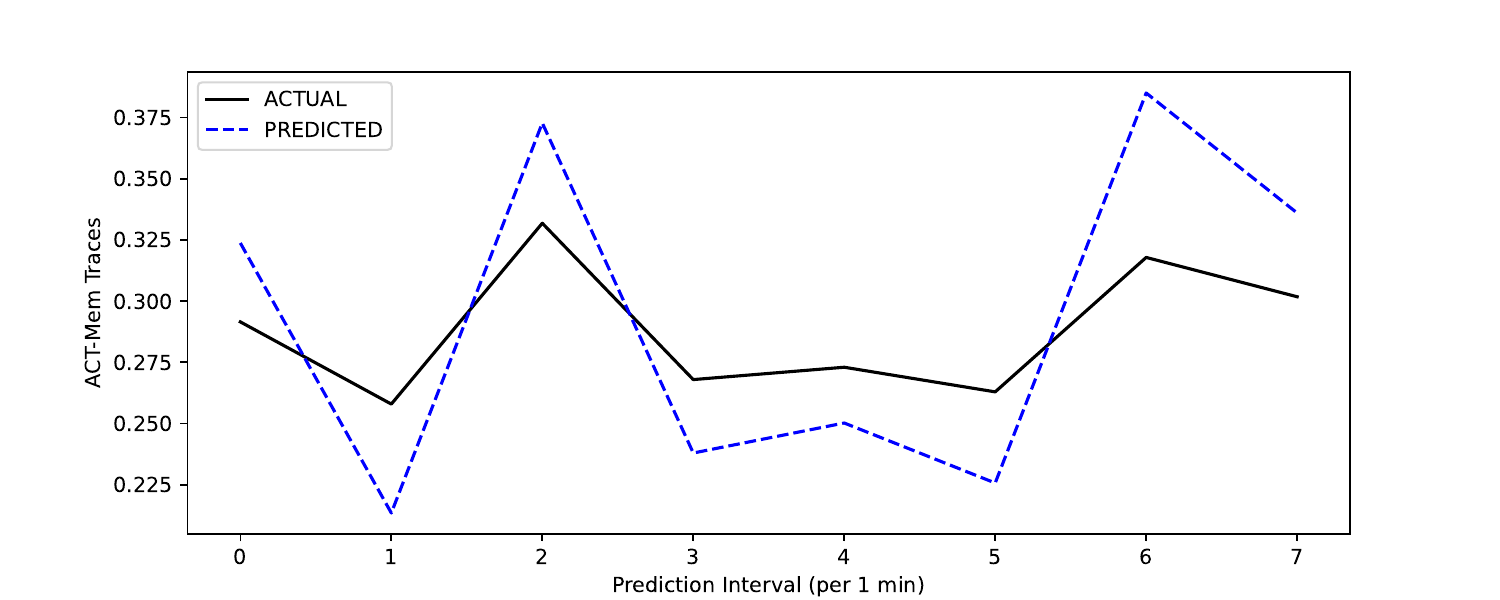}};
		\node[black] (B) at ($(A.south)!-0.05!(A.north)$) {\footnotesize  Prediction Interval (day)};
		\node[black,rotate=90] (C) at ($(A.west)!-0.03!(A.east)$) {\footnotesize  Memory Utilization};
	\end{tikzpicture}
	\caption{1 day}
\end{subfigure}
\caption{Actual versus predicted workload for ACT-Mem}
\label{res-fig:1d}
\end{figure*}

\begin{table*}[!htbp]
\caption{Prediction accuracy in terms of various error metrics} \label{res-tab:2}
\centering
\resizebox{0.95\textwidth}{!}{
	\begin{tabular}{c|l|lll|lll|lll}
		\hline
		\multirow{2}{*}{\textbf{DT}} & \textbf{PI} & \multicolumn{3}{c|}{\textbf{RMSE}} &\multicolumn{3}{c|}{\textbf{MAE}} &\multicolumn{3}{c}{\textbf{MAPE}} \\
		\cline{3-11} &\textbf{(minutes)}&\textbf{Training} & {\textbf{Testing}} & {\textbf{Time (ms)}}  & \textbf{Training} & {\textbf{Testing}} & {\textbf{Time (ms)}} & \textbf{Training} &{\textbf{Testing}} & {\textbf{Time (ms)}} \\ \hline
		\multirow{6}{*}{\rotatebox{90}{GCD-Mem}} &5& 0.03244&	0.03326	 & 31327 &	0.01414	&0.01474& 312095 &	0.62621 &	0.63400	&311897 \\ 
		& 10	&0.03454&	0.03110	&158506 &	0.03017&	0.03013 &157200&	0.63237	&0.63645 &	158061 \\
		& 20	&0.07278	&0.07960	&80281	&0.04007	&0.04004	&79846	&0.62473	&0.67289 &79467 \\
		&30	&0.09899	&0.11478	&53279	&0.06259	&0.06257	&53036	&0.44932	&0.44450	& 53130 \\
		&60	&0.10425	&0.10424	&27003	&0.07014	&0.07014	&27145	&0.45732	&0.50127	&26781 \\
		&1440	&0.10573	&0.06833	&1175	&0.04489	&0.12574	&1188	&0.33195	&0.34059	&1195 \\ \hline	
		\multirow{6}{*}{\rotatebox{90}{ACT-CPU}} &5&  0.11002	& 0.11141	& 85825	& 0.08520	& 0.08495	& 86164	& 0.28381	& 0.28523	& 84143\\ 
		& 10	& 0.11417	&0.11138	&43530	&0.08354	&0.08921	&43590	&0.28200	&0.28646	&43188\\
		& 20	&0.12640	&0.13742	&21880	&0.10018	&0.10105	&21883	&0.29043	&0.29983	&21887 \\
		&30	&	0.13346	&0.12462	&14666	&0.10465	&0.09378	&14626	&0.28780	&0.28347	&14345 \\
		&60	& 0.13842	&0.13490	&7380	&0.10709	&0.10831	&7393	&0.28540	&0.28763	&7193\\
		&1440	&0.12842	&0.12342	&2994	&0.10302	&0.10403	&2333	&0.27302	&0.27493 &2013 \\ \hline	
		\multirow{6}{*}{\rotatebox{90}{ACT-Mem}} &5& 0.09001	&0.09048	&85977	&0.07476	&0.07082	&85701	&0.28381	&0.28523	&85341 \\ 
		& 10	&0.13681	&0.13032	&43381	&0.10270	&0.09797	&43391	&0.28200	&0.28646	&43205 \\
		& 20	& 0.14212	&0.14148	&21829	&0.11325	&0.09517	&21929	&0.29043	&0.29983	&20985\\
		&30	& 0.16532	&0.15175	&14687	&0.11931	&0.12562	&14631	&0.28780	&0.28347 & 14032\\
		&60	& 0.16293	& 0.17491	& 7494	& 0.12297	& 0.14509	& 7359	& 0.28540	& 0.28763	& 7111\\
		&1440	&0.16343	&0.16953	&2347	&0.12846	&0.13866	&2235	&0.27302	&0.27493 & 2013 \\ \hline	
\end{tabular}}
\\ \footnotesize{\scriptsize DT: Workload Data Traces, GCD-Mem: Google Cluster Memory Traces, ACT-CPU: Alibaba Cluster CPU Traces, ACT-Mem: Alibaba Clsuter Memory Traces, ms: mili-seconds}
\end{table*}
The performance of the CA-QNN prediction model is thoroughly evaluated across a diverse range of heterogeneous cloud applications and varying VM resource utilization. Fig. \ref{res-fig:1a} illustrates the comparison between actual and predicted values for  GCD-CPU workloads for PI of five minutes (Fig. \ref{res-fig:1a}(a)), one hour (Fig. \ref{res-fig:1a}(b)), and one day (Fig. \ref{res-fig:1a}(c)).  Additionally, Fig. \ref{res-fig:1b}, Fig. \ref{res-fig:1c}, and Fig. \ref{res-fig:1d} provide a similar comparison for GCD-Mem, ACT-CPU, and ACT-Mem workloads, respectively. These diverse range of results indicate that the proposed method accurately predicts future values for all types of workloads for large scale as well as small scale prediction intervals, closely matching the actual values. The CA-QNN model effectively captures sudden spikes and drops in workload demands, demonstrating its robustness and precision in handling dynamic resource utilization patterns. 

Fig. \ref{res-figbest-cpu} and Fig. \ref{res-figbest-mem} presents the intermediate values of four randomly selected VQNN network vectors ($\mathcal{A}1$, $\mathcal{A}2$, $\mathcal{A}3$, $\mathcal{A}4$) with optimal fitness scores ($\digamma (\uplus_{Opm})$ \textit{score}) across 50 epochs for all four workload data traces. These three-dimensional graphs illustrate the evolving status of the best candidates, guiding the VQNN towards minimizing prediction error from 10 to 50 epochs, and ultimately producing a highly accurate CA-QNN prediction model.

\begin{figure*}[!htbp]
\centering
\begin{subfigure}[t]{0.49\textwidth}
	\centering
	\begin{tikzpicture}[scale=1, transform shape]
		\pgfmathsetmacro{\gconv}{63.85498}
		\begin{axis}[
			view={110}{20},
			width=9cm,
			height=6cm,
			grid=major,
			xmin=0,xmax=4.5,
			ymin=0,ymax=5.5,
			zmin=0.0,
			xtick={1,2,3,4},
			xticklabels={$\mathcal{A}1$, $\mathcal{A}2$, $\mathcal{A}3$, $\mathcal{A}4$},xticklabel style={font={\scriptsize}},
			ytick={1,2,3,4,5},
			yticklabels={10, 20, 30, 40, 50},yticklabel style={font={\scriptsize}},
			xlabel={\scriptsize \rotatebox{50}{\textit{Architectures}}},
			zlabel={\scriptsize  \textit{Optimal} ($\uplus_{Opm}$) \textit{Solution}},
			ylabel={\scriptsize \textit{Epochs}},
			axis equal,
			]
			\path let \p1=($(axis cs:0,0,1)-(axis cs:0,0,0)$) in 
			\pgfextra{\pgfmathsetmacro{\conv}{2*\y1}
				\ifx\gconv\conv
				\typeout{z-scale\space good!}
				\else
				\typeout{Kindly\space consider\space setting\space the\space 
					prefactor\space of\space z\space to\space \conv}
				\fi     
			};  
			
			\pgfplotsset{3d bars/.style={only marks,scatter,mark=half cube*,mark size=0.3cm, 
					3d cube color=#1,point meta=0,
					,visualization depends on={\gconv*z \as \myz},
					scatter/@pre marker code/.append style={/pgfplots/cube/size z=\myz},}}
			
			\addplot3[3d bars=blue!40]
			coordinates {(1,1,0.8520)(1,2,0.8319)(1,3,0.819)(1,4,0.8655)(1,5,0.756551)};
			
			\addplot3[3d bars=red!40]
			coordinates {(2,1,1.010561)(2,2,0.970810)(2,3,0.957113)(2,4,0.857113)(2,5,0.65403)};
			
			\addplot3[3d bars=violet!40]
			coordinates {(3,1,0.972086)(3,2,0.962789)(3,3,0.941070)(3,4,0.832788)(3,5,0.731944)};
			
			\addplot3[3d bars=teal!40]
			coordinates {(4,1,0.979647)(4,2,0.978615)(4,3,0.872217)(4,4,0.749038)(4,5,0.647967)};
			
		\end{axis}
	\end{tikzpicture}
	\caption{GCD-CPU}
\end{subfigure}
\hfill
\begin{subfigure}[t]{0.49\textwidth}
	\centering
	\begin{tikzpicture}[scale=1, transform shape]
		\pgfmathsetmacro{\gconv}{63.85498}
		\begin{axis}[
			view={110}{20},
			width=9cm,
			height=6cm,
			grid=major,
			xmin=0,xmax=4.5,
			ymin=0,ymax=5.5,
			zmin=0.0,
			xtick={1,2,3,4},
			xticklabels={$\mathcal{A}1$, $\mathcal{A}2$, $\mathcal{A}3$, $\mathcal{A}4$},xticklabel style={font={\scriptsize}},
			ytick={1,2,3,4,5},
			yticklabels={10, 20, 30, 40, 50},yticklabel style={font={\scriptsize}},
			xlabel={\scriptsize \rotatebox{50}{\textit{Architectures}}},
			zlabel={\scriptsize  \textit{Optimal} ($\uplus_{Opm}$) \textit{Solution}},
			ylabel={\scriptsize \textit{Epochs}},
			axis equal,
			]
			\path let \p1=($(axis cs:0,0,1)-(axis cs:0,0,0)$) in 
			\pgfextra{\pgfmathsetmacro{\conv}{2*\y1}
				\ifx\gconv\conv
				\typeout{z-scale\space good!}
				\else
				\typeout{Kindly\space consider\space setting\space the\space 
					prefactor\space of\space z\space to\space \conv}
				\fi     
			};  
			
			\pgfplotsset{3d bars/.style={only marks,scatter,mark=half cube*,mark size=0.3cm, 
					3d cube color=#1,point meta=0,
					,visualization depends on={\gconv*z \as \myz},
					scatter/@pre marker code/.append style={/pgfplots/cube/size z=\myz},}}
			
			\addplot3[3d bars=blue!40]
			coordinates {(1,1,1.344357)(1,2,1.265275)(1,3,1.256596)(1,4,1.224916)(1,5,1.175423)};
			
			\addplot3[3d bars=red!40]
			coordinates {(2,1,1.302019)(2,2,1.256703)(2,3,1.166835)(2,4,1.148698)(2,5,1.145564)};
			
			\addplot3[3d bars=violet!40]
			coordinates {(3,1,1.232692)(3,2,1.193099)(3,3,1.162051)(3,4,1.161795)(3,5,1.130079)};
			
			\addplot3[3d bars=teal!40]
			coordinates {(4,1,1.288520)(4,2,1.248797)(4,3,1.248797)(4,4,1.154878)(4,5,1.142424)};
			
		\end{axis}
	\end{tikzpicture}
	\caption{ACT-CPU}
\end{subfigure}
\caption{ Comprehensive architectural learning-based CA-QNN optimization over epochs for CPU workload traces}
\label{res-figbest-cpu}
\end{figure*}
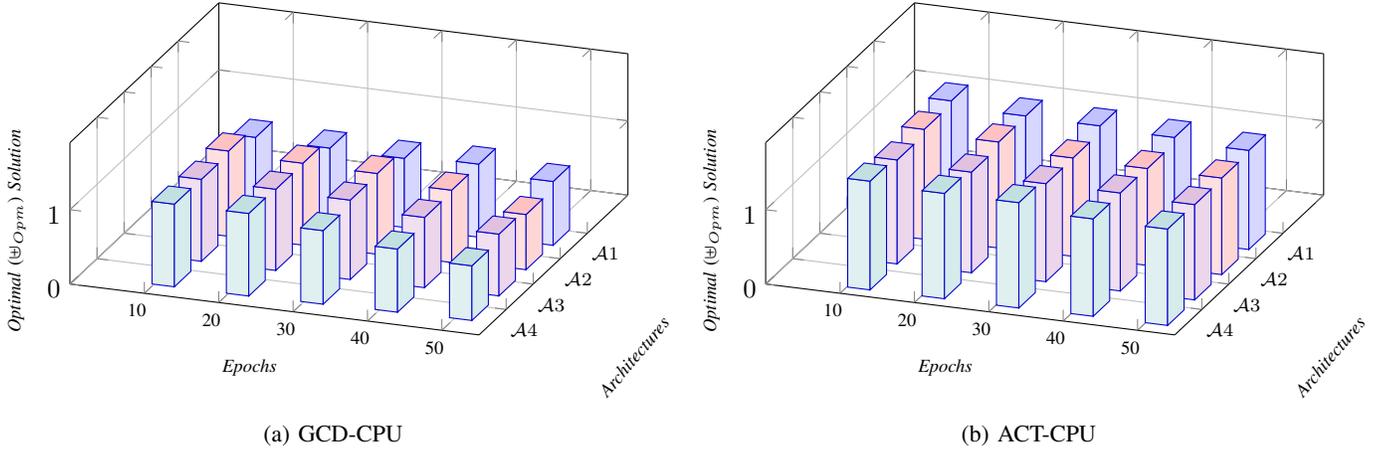
\begin{figure*}[!htbp]
\centering
\begin{subfigure}[t]{0.49\textwidth}
	\centering
	\begin{tikzpicture}[scale=1, transform shape]
		\pgfmathsetmacro{\gconv}{63.85498}
		\begin{axis}[
			view={110}{20},
			width=9cm,
			height=6cm,
			grid=major,
			xmin=0,xmax=4.5,
			ymin=0,ymax=5.5,
			zmin=0.0,
			xtick={1,2,3,4},
			xticklabels={$\mathcal{A}1$, $\mathcal{A}2$, $\mathcal{A}3$, $\mathcal{A}4$},xticklabel style={font={\scriptsize}},
			ytick={1,2,3,4,5},
			yticklabels={10, 20, 30, 40, 50},yticklabel style={font={\scriptsize}},
			xlabel={\scriptsize \rotatebox{50}{\textit{Architecture}}},
			zlabel={\scriptsize  \textit{Optimal} ($\uplus_{Opm}$) \textit{Solution}},
			ylabel={\scriptsize \textit{Epochs}},
			axis equal,
			]
			\path let \p1=($(axis cs:0,0,1)-(axis cs:0,0,0)$) in 
			\pgfextra{\pgfmathsetmacro{\conv}{2*\y1}
				\ifx\gconv\conv
				\typeout{z-scale\space good!}
				\else
				\typeout{Kindly\space consider\space setting\space the\space 
					prefactor\space of\space z\space to\space \conv}
				\fi     
			};  
			
			\pgfplotsset{3d bars/.style={only marks,scatter,mark=half cube*,mark size=0.3cm, 
					3d cube color=#1,point meta=0,
					,visualization depends on={\gconv*z \as \myz},
					scatter/@pre marker code/.append style={/pgfplots/cube/size z=\myz},}}
			
			\addplot3[3d bars=blue!40]
			coordinates {(1,1,0.382063)(1,2,0.366700)(1,3,0.363546)(1,4,0.360244)(1,5,0.363282)};
			
			\addplot3[3d bars=red!40]
			coordinates {(2,1,0.371900)(2,2,0.366715)(2,3,0.360917)(2,4,0.355360)(2,5,0.359062)};
			
			\addplot3[3d bars=violet!40]
			coordinates {(3,1,0.371254)(3,2,0.366031)(3,3,0.362875)(3,4,0.358256)(3,5,0.355159)};
			
			\addplot3[3d bars=teal!40]
			coordinates {(4,1,0.368428)(4,2,0.365191)(4,3,0.362346)(4,4,0.359543)(4,5,0.355458)};
			
		\end{axis}
	\end{tikzpicture}
	\caption{GCD-Mem }
\end{subfigure}
\hfill
\begin{subfigure}[t]{0.49\textwidth}
	\centering
	\begin{tikzpicture}[scale=1, transform shape]
		\pgfmathsetmacro{\gconv}{63.85498}
		\begin{axis}[
			view={110}{20},
			width=9cm,
			height=6cm,
			grid=major,
			xmin=0,xmax=4.5,
			ymin=0,ymax=5.5,
			zmin=0.0,
			xtick={1,2,3,4},
			xticklabels={$\mathcal{A}1$, $\mathcal{A}2$, $\mathcal{A}3$, $\mathcal{A}4$},xticklabel style={font={\scriptsize}},
			ytick={1,2,3,4,5},
			yticklabels={10, 20, 30, 40, 50},yticklabel style={font={\scriptsize}},
			xlabel={\scriptsize \rotatebox{50}{\textit{Architecture}}},
			zlabel={\scriptsize  \textit{Optimal} ($\uplus_{Opm}$) \textit{Solution}},
			ylabel={\scriptsize \textit{Epochs}},
			axis equal,
			]
			\path let \p1=($(axis cs:0,0,1)-(axis cs:0,0,0)$) in 
			\pgfextra{\pgfmathsetmacro{\conv}{2*\y1}
				\ifx\gconv\conv
				\typeout{z-scale\space good!}
				\else
				\typeout{Kindly\space consider\space setting\space the\space 
					prefactor\space of\space z\space to\space \conv}
				\fi     
			};  
			
			\pgfplotsset{3d bars/.style={only marks,scatter,mark=half cube*,mark size=0.3cm, 
					3d cube color=#1,point meta=0,
					,visualization depends on={\gconv*z \as \myz},
					scatter/@pre marker code/.append style={/pgfplots/cube/size z=\myz},}}
			
			\addplot3[3d bars=blue!40]
			coordinates {(1,1,1.007433)(1,2,0.992347)(1,3,0.972020)(1,4,0.963766)(1,5,0.963766)};
			
			\addplot3[3d bars=red!40]
			coordinates {(2,1,0.967324)(2,2,0.967324)(2,3,0.965786)(2,4,0.916872)(2,5,0.913502)};
			
			\addplot3[3d bars=violet!40]
			coordinates {(3,1,0.959861)(3,2,0.929516)(3,3,0.909568)(3,4,0.897086)(3,5,0.891272)};
			
			\addplot3[3d bars=teal!40]
			coordinates {(4,1,0.988710)(4,2,0.968720)(4,3,0.968720)(4,4,0.940335)(4,5,0.940335)};
			
		\end{axis}
	\end{tikzpicture}
	\caption{ACT-Mem}
\end{subfigure}
\caption{Comprehensive architectural learning-based CA-QNN optimization over epochs for Memory workload traces}
\label{res-figbest-mem}
\end{figure*}
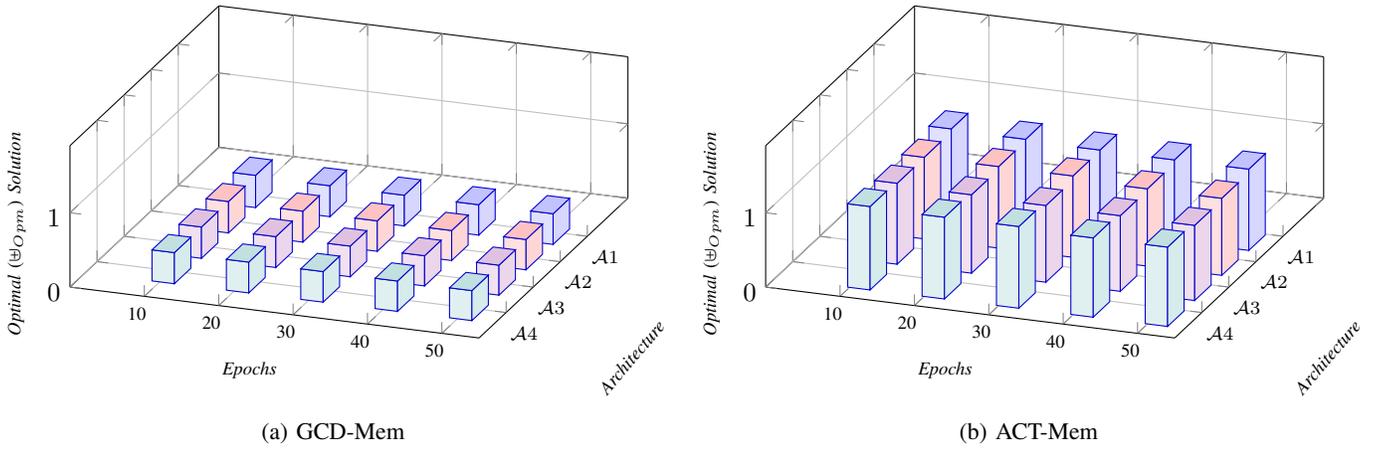

\subsection{Comparison}
The performance of CA-QNN prediction model is compared with neural network optimized by Backpropagation (BPNN) algorithm \cite{lu2016rvlbpnn}, SaDE algorithm \cite{kumar2018workload}, BaDE algorithm \cite{kumar2018workload}, LSTM-RNN \cite{kumar2018long}, and EQNN \cite{singh2021quantum}, CLIN \cite{kim2020forecasting}, MCT-AQNN \cite{gupta2024multiple} workload prediction models. The concise description of these approaches is given in the related work (Section \ref{rw}). The comparison is analysed in terms of RMSE, MSE, MAE, \textit{Normalised RMSE}, \textit{Average training time}, \textit{Average number of epochs}, and \textit{Computational complexity} over wide range of PI from 5 minutes to 60 minutes using GCD-CPU and GCD-Mem workload data traces. 

\subsubsection{Accuracy metrics}
The resultant values of accuracy metrics (measured in terms of prediction  errors) including RMSE, MAE, and MAPE of CA-QNN are compared with all seven state-of-the-art methods in Table \ref{tab:rmse}, Table \ref{tab:mae}, and Table \ref{tab:mape}, respectively.
The CA-QNN model demonstrates a reduction in RMSE by 95.87\%, 92.67\%, 88.28\%, 88.02\%, 93.40\%, 91.27\%, 8.88\% compared to BPNN, SaDE, BaDE, CLIN, LSTM, EQNN, and MCT-AQNN, respectively, as shown in Table \ref{tab:rmse}. Additionally, Table \ref{tab:mae} reports enhanced accuracy through a reduction in MAE, where CA-QNN outperforms BPNN, SaDE, BaDE, CLIN, LSTM, EQNN, and MCT-AQNN by 95.30\%, 94.59\%, 85.30\%, 81.15\%, 91.50\%, 58.70\%, 6.36\%, respectively. The MAPE values are also decreased by 77.70\%, 78.44\%, 75.05\%, 50.93\%, 68.24\%, 25.38\%, 7.88\% over BPNN, SaDE, BaDE, CLIN, LSTM, EQNN, and MCT-AQNN, respectively, as observed in Table \ref{tab:mape}. Therefore, CA-QNN has outperformed existing models for dynamic and broad-range cloud workload prediction due to its superior prediction accuracy achieved through comprehensive architectural and parametric learning and optimization.


\begin{table*}[!htbp]
\caption{RMSE ($ \times 10^{-3}$): CA-QNN versus state-of-the-arts } \label{tab:rmse}
\centering
\resizebox{0.9\textwidth}{!}{
\begin{tabular}{cccccccccc}
\hline
\textbf{DT} & \textbf{ PI} (minute)& \textbf{ BPNN}\cite{lu2016rvlbpnn} &\textbf{SaDE} \cite{kumar2018workload}&\textbf{BaDE}\cite{kumar2020biphase}  & \textbf{CLIN}\cite{kim2020forecasting} &\textbf{LSTM-RNN} \cite{kumar2018long} &\textbf{EQNN}\cite{singh2021quantum} &\textbf{MCT-AQNN} \cite{gupta2024multiple}&\textbf{CA-QNN}\\ \hline
\multirow{4}{*}{\rotatebox{90}{GCPU}} &5& 18.2& 12.7& 10.4 &7.78 &19.7 &9.87&  \underline{5.97}&  \textbf{4.24}\\ 
& 10 & 21.9& 20.1& 10.9 &7.34 &13.2 &17.5&  \textbf{1.96}&  \underline{4.24}\\ 
&30	& 40.1 &35.6 & 31.1 &16.6 & 13.4&31.7&  \underline{11.6} &  \textbf{11.3}\\ 
&60	& 42.9& 38.0& 33.6 &21.7 &10.2 &20.93&  \underline{10.1}&  \textbf{10.02} \\ \hline
\multirow{4}{*}{\rotatebox{90}{GMem}} &5& 19.9 &11.2 &7.00& 6.85 & 19.7& 9.40 &  \underline{0.90} &  \textbf{0.82} \\ 
& 10 &37.3&	14.2	&10.7 &	7.89&	19.1 &10.5 &	 \underline{3.73}	& \textbf{3.11} \\
&30	&54.3	&39.4	&37.6	&15.0	&19.4	&48.5	&  \underline{9.90}	&  \textbf{9.89}\\
&60	&62.6	&48.2	&40.6	&19.6	&11.4	&39.3	 &  \underline{10.9}	&  \textbf{10.4}\\ \hline	
\end{tabular}}
\\ \footnotesize{\scriptsize DT: Workload Data Traces, GCPU: GCD-CPU,GMem: GCD-Mem}
\end{table*}
\begin{table*}[!htbp]
\caption{MAE ($ \times 10^{-2}$): CA-QNN versus state-of-the-arts} \label{tab:mae}
\centering
\resizebox{0.9\textwidth}{!}{
\begin{tabular}{cccccccccc}
\hline
\textbf{DT} & \textbf{ PI} (minutes)& \textbf{ BPNN}\cite{lu2016rvlbpnn} &\textbf{SaDE} \cite{kumar2018workload}&\textbf{BaDE}\cite{kumar2020biphase}  & \textbf{CLIN}\cite{kim2020forecasting} &\textbf{LSTM-RNN} \cite{kumar2018long} &\textbf{EQNN}\cite{singh2021quantum} &\textbf{MCT-AQNN} \cite{gupta2024multiple} &\textbf{CA-QNN} \\ \hline
\multirow{4}{*}{\rotatebox{90}{GCPU}} &5&36.8 &59.7 &24.9&7.12&27.6 & 7.34 & \underline{4.57}& \textbf{4.47} \\ 
& 10 &27.5 &25.1 &12.6 &7.86 & 14.8& 3.42 & \underline{1.98} & \textbf{1.61}\\ 
&30 &36.3 &13.7 &11.7 &9.11 &14.2 & 10.63&  \underline{7.51} & \textbf{7.47}	\\ 
&60	&34.2 &15.6 &14.7 &10.3 &14.9 & 7.89 &  \underline{7.25} &  \textbf{7.23} \\ \hline
\multirow{4}{*}{\rotatebox{90}{GMem}} &5 &31.3 &27.2 &10.0 &7.80 &17.3 & 3.56&  \underline{1.57} & \textbf{1.47}  \\ 
& 10 &38.4 &31.9 &11.6 &7.74 &21.5 &4.43& \underline{3.05} & \textbf{3.01} \\
&30	&32.3 &16.4 &13.6 &8.36 &19.4 & 7.74& \underline{6.54} &  \textbf{6.25} \\
&60	&33.6 &16.5 &14.6 &9.15 &19.6 &9.34 & \underline{7.02} & \textbf{7.01} \\ \hline	
\end{tabular}}
\\  \footnotesize{\scriptsize DT: Workload Data Traces, GCPU: GCD-CPU,GMem: GCD-Mem}
\end{table*}
\begin{table*}[!htbp]
\caption{MAPE ($ \times 10^{-1}$): CA-QNN versus state-of-the-arts} \label{tab:mape}
\centering
\resizebox{0.9\textwidth}{!}{
\begin{tabular}{cccccccccc}
\hline
\textbf{DT} & \textbf{ PI} (minutes)& \textbf{ BPNN}\cite{lu2016rvlbpnn} &\textbf{SaDE} \cite{kumar2018workload}&\textbf{BaDE}\cite{kumar2020biphase}  & \textbf{CLIN}\cite{kim2020forecasting} &\textbf{LSTM-RNN} \cite{kumar2018long} &\textbf{EQNN}\cite{singh2021quantum} &\textbf{MCT-AQNN} \cite{gupta2024multiple}&\textbf{CA-QNN}\\ \hline
		\multirow{4}{*}{\rotatebox{90}{GCPU}} &5&25.3 &20.4 &15.5 &17.1 &17.4 &6.34 & \underline{5.88} &  \textbf{5.38}  \\ 
		& 10 &18.3 & 15.3 &12.9 &10.3 &16.4 &  \underline{7.89}&9.43 &  \textbf{6.73}\\ 
		&30 &17.8 &9.58 &8.83 &6.65 &12.5 &5.32 & \underline{4.27} & \textbf{3.97}	\\ 
		&60	&18.4 &15.1 &14.2 &9.57 &16.2 & 6.34& \underline{4.70} & \textbf{4.55} \\ \hline
		\multirow{4}{*}{\rotatebox{90}{GMem}} &5&23.9 &19.4 &18.9 &7.34 &14.3 & \underline{6.45} &6.75 & \textbf{6.34}  \\ 
		& 10 &19.4 &20.1 &18.3 &13.6 &13.4 &  \underline{8.05}&11.9 &  \textbf{6.36} \\
		&30	&20.6 &20.9 &17.8 &9.05 &14.7 &7.31 & \underline{4.82} &  \textbf{4.44} \\
		&60	&20.3 &23.5 &19.6 &14.7 &14.3 &9.42 &  \underline{6.72} &  \textbf{5.01}\\ \hline	
\end{tabular}}
\\  \footnotesize{\scriptsize DT: Workload Data Traces, GCPU: GCD-CPU,GMem: GCD-Mem}
\end{table*}

\subsubsection{Normalized RMSE}
\begin{figure*} [!htbp]
\centering
\begin{tikzpicture}
	\begin{axis}[
		width=0.9\textwidth,
		height=6.0cm,
		ybar,
		ymin=0,
		ymax=6.2,
		ytick={1,2,3,4,5,6},yticklabel style={font={\scriptsize}},
		ybar=5pt,
		bar width=7pt,
		enlarge x limits=0.15,
		legend style={at={(0.5,1.15)},
			anchor=north,legend columns=-1},
		ylabel={\textit{Normalized RMSE}},
		xlabel={\textit{ Prediction Interval (minutes)}},
		symbolic x coords={G1, 5, 10,  30,  60, G5},
		xtick=data,xticklabel style={font={\scriptsize}},
		nodes near coords,
		every node near coord/.append style={rotate=90, anchor=west, font={\scriptsize}},
		ymajorgrids=true,
		xmajorgrids=true,
		grid style=dashed, thin
		]
		\addplot[draw=magenta,thick, fill=yellow!10!] coordinates{(5, 4.29) (10, 5.16) (30, 3.54) (60, 4.28)}; \addlegendentry{\scriptsize  BPNN} \cite{lu2016rvlbpnn}
		\addplot[fill=white, thick, draw=teal, pattern color=black!30, postaction={
			pattern=grid
		}] coordinates{(5, 2.99) (10, 4.74) (30, 3.15) (60, 3.79)}; \addlegendentry{\scriptsize SaDE \cite{kumar2018workload}}
		\addplot+[fill=white,thick, pattern color=black!30, draw=olive, postaction={
			pattern=dots
		}] coordinates{(5, 2.45) (10, 2.57) (30, 2.75) (60, 3.35)}; \addlegendentry{\scriptsize BaDE \cite{kumar2020biphase}}
		\addplot+[fill=white, thick, pattern color=black!30, draw=magenta, postaction={
			pattern=vertical lines
		}] coordinates{(5, 1.83) (10, 1.73) (30, 1.92) (60, 2.16)};\addlegendentry{\scriptsize CLIN \cite{kim2020forecasting}}
		\addplot[fill=white,thick, draw=cyan,pattern color=black!30, postaction={
			pattern=north east lines
		}] coordinates{(5, 4.64) (10, 3.11) (30, 1.18) (60, 1.02)}; \addlegendentry{\scriptsize LSTM-RNN \cite{kumar2018long}}
		\addplot[fill=white, thick, pattern color=black!30, draw=orange!80!,  postaction={
			pattern=horizontal lines
		}] coordinates{(5, 1.32) (10, 1.12) (30, 1.80) (60, 1.08)}; \addlegendentry{\scriptsize EQNN \cite{singh2021quantum}}
		\addplot[fill=violet!10, thick, pattern color=violet!30, draw=violet!80!,  postaction={
			pattern=none
		}] coordinates{(5, 1.40) (10, 1.46) (30, 1.03) (60, 1.01)}; \addlegendentry{\scriptsize MCT-AQNN \cite{gupta2024multiple}}
		\addplot[red,thick,line legend,sharp plot,nodes near coords={},
		update limits=false,shorten >=-3mm,shorten <=-3mm]
		coordinates {(G1, 1) (G5, 1)}
		node[midway,above,font=\bfseries\sffamily, style={rotate=90, anchor=west}]{\scriptsize CA-QNN (1.00)};
	\end{axis}
\end{tikzpicture}
\caption{Normalized RMSE with GCD-CPU} \label{fig:N-cpu}
\end{figure*}
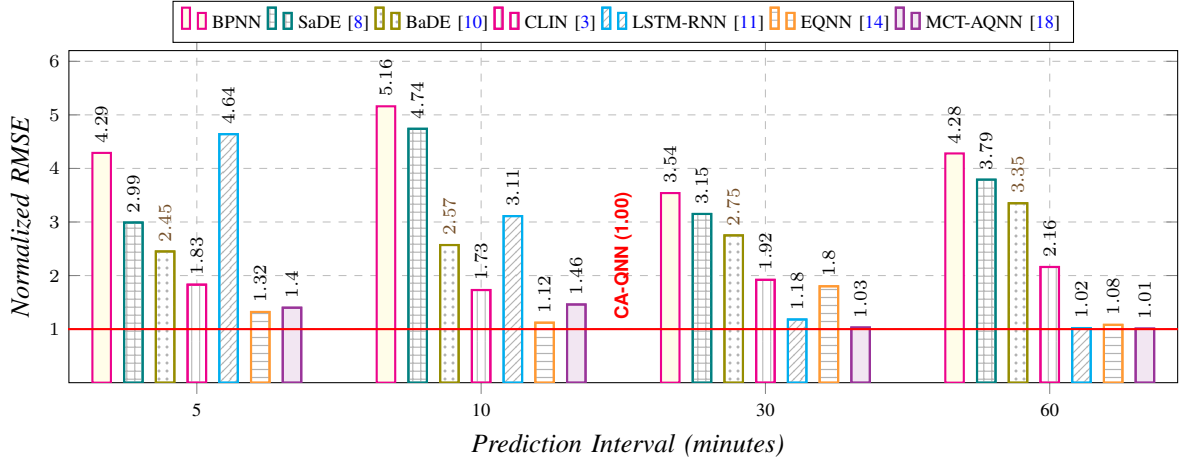

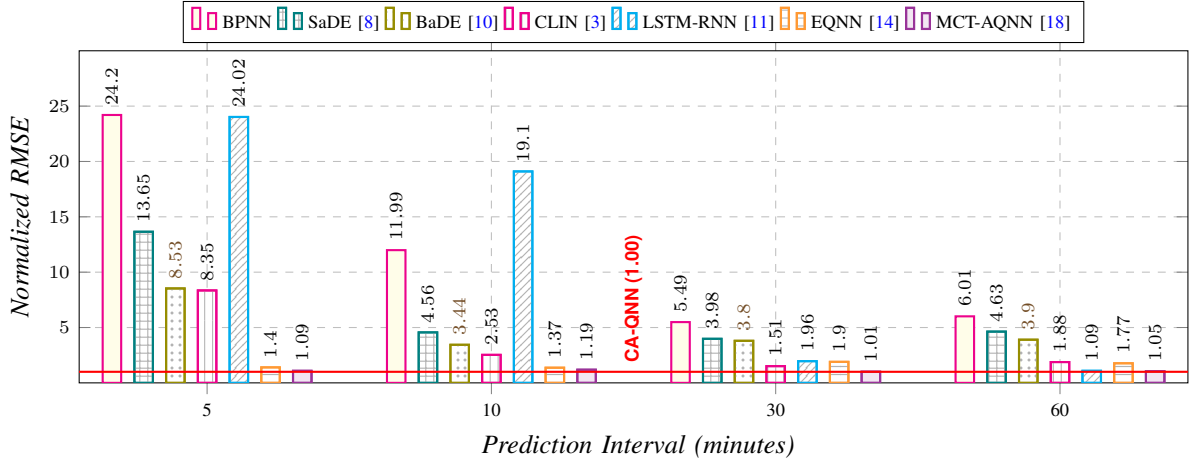
\begin{figure*} [!htbp]
\centering
\begin{tikzpicture}
	\begin{axis}[
		width=0.9\textwidth,
		height=6.0cm,
		ybar,
		ymin=0,
		ymax=30,
		ytick={5,10,15,20,25},yticklabel style={font={\scriptsize}},
		ybar=5pt,
		bar width=7pt,
		enlarge x limits=0.15,
		legend style={at={(0.5,1.15)},
			anchor=north,legend columns=-1},
		ylabel={\textit{Normalized RMSE}},
		xlabel={\textit{ Prediction Interval (minutes)}},
		symbolic x coords={G1, 5, 10,  30,  60, G5},
		xtick=data,xticklabel style={font={\scriptsize}},
		nodes near coords,
		every node near coord/.append style={rotate=90, anchor=west,font={\scriptsize}},
		ymajorgrids=true,
		xmajorgrids=true,
		grid style=dashed, thin
		]
		\addplot[draw=magenta,thick, fill=yellow!10!] coordinates{(5, 24.2) (10, 11.99) (30, 5.49) (60, 6.01)}; \addlegendentry{\scriptsize  BPNN} \cite{lu2016rvlbpnn}
		\addplot[fill=white, thick, draw=teal, pattern color=black!30, postaction={
			pattern=grid
		}] coordinates{(5, 13.65) (10, 4.56) (30, 3.98) (60, 4.63)}; \addlegendentry{\scriptsize SaDE \cite{kumar2018workload}}
		\addplot+[fill=white,thick, pattern color=black!30, draw=olive, postaction={
			pattern=dots
		}] coordinates{(5, 8.53) (10, 3.44) (30, 3.80) (60, 3.90)}; \addlegendentry{\scriptsize BaDE \cite{kumar2020biphase}}
		\addplot+[fill=white, thick, pattern color=black!30, draw=magenta, postaction={
			pattern=vertical lines
		}] coordinates{(5, 8.35) (10, 2.53) (30, 1.51) (60, 1.88)};\addlegendentry{\scriptsize CLIN \cite{kim2020forecasting}}
		\addplot[fill=white,thick, draw=cyan,pattern color=black!30, postaction={
			pattern=north east lines
		}] coordinates{(5, 24.02) (10, 19.1) (30, 1.96) (60, 1.09)}; \addlegendentry{\scriptsize LSTM-RNN \cite{kumar2018long}}
		\addplot[fill=white, thick, pattern color=black!30, draw=orange!80!,  postaction={
			pattern=horizontal lines
		}] coordinates{(5, 1.4) (10, 1.37) (30, 1.90) (60, 1.77)}; \addlegendentry{\scriptsize EQNN \cite{singh2021quantum}}
		\addplot[fill=violet!10, thick, pattern color=violet!30, draw=violet!80!,  postaction={
			pattern=none}] coordinates{(5, 1.09) (10, 1.19) (30, 1.01) (60, 1.05)}; \addlegendentry{\scriptsize MCT-AQNN \cite{gupta2024multiple}}
		\addplot[red,thick,line legend,sharp plot,nodes near coords={},
		update limits=false,shorten >=-3mm,shorten <=-3mm]
		coordinates {(G1, 1) (G5, 1)}
		node[midway,above,font=\bfseries\sffamily, style={rotate=90, anchor=west}]{\scriptsize CA-QNN (1.00)};
	\end{axis}
\end{tikzpicture}
\caption{Normalized RMSE with GCD-Mem} \label{fig:N-mem}
\end{figure*}

Fig. \ref{fig:N-cpu}  and Fig. \ref{fig:N-mem} compare the normalised RMSE results  obtained for CA-QNN model with the relative RMSE of seven comparative approaches for GCD-CPU and GCD-Mem workloads traces, respectively. A value of $1.00$ represents the CA-QNN model's performance, with higher values indicating worse results. RMSE values from other prediction methods are normalized relative to the  RMSE values of CA-QNN. The majority of the prediction error reduction trend is: BPNN $<$ SaDE $<$ BaDE $<$ LSTM $<$ CLIN $<$ EQNN $<$ MCT-AQNN $<$ CA-QNN for both data traces. BPNN iteratively minimizes error using a single solution, while SaDE, BaDE, EQNN, and MCT-AQNN employ evolutionary algorithms for multi-dimensional exploration and optimization. BaDE's dual adaptation surpasses SaDE's capabilities, and while LSTM achieves effective accuracy, it is prone to overfitting due to random weight initialization. CA-QNN, MCT-AQNN, and EQNN use Qubits population and evolutionary optimization for better pattern learning. CA-QNN excels with comprehensive architectural optimization, extensive exploration, and multi-dimensional structural and hyper-parameter tuning, leading to more accurate predictions than EQNN and MCT-AQNN. 

\subsubsection{Optimization Analysis}

\begin{figure*}[!htbp]
\centering
\begin{subfigure}[t]{0.49\textwidth}
	\centering
	\begin{tikzpicture}
		\begin{axis}[
			width=.99\columnwidth,
			height=6.0cm,
			ymin=0,
			ymax=300,ytick={0,50,100,150,200,250},yticklabel style={font={\scriptsize}},
			xtick={1,2,3,4}, xticklabels={5,10,30,60},xticklabel style={font={\scriptsize}},
			ybar,
			ymajorgrids=true,
			xmajorgrids=true,
			grid style=dashdotted,
			bar width=6.0pt,
			enlarge x limits=0.12,
			xlabel={\textit{ Prediction Interval (minutes)}},
			nodes near coords,
			every node near coord/.append style={rotate=90, anchor=west, color=black,font={\scriptsize}},
			legend style={at={(0.55,1.2)},
				anchor=north,legend columns=5},
			ylabel={\textit{Training Time (seconds)}}]
			
			\addplot+[ybar,fill=white, thick, draw=teal, pattern color=black!30, postaction={
				pattern=grid,thin}]plot coordinates{ (1,48.54)(2,31.7)(3,5.51)(4,3.5)}; \addlegendentry{\scriptsize SaDE \cite{kumar2018workload}}
			\addplot+[ybar,fill=white,thick, pattern color=black!30, draw=olive, postaction={
				pattern=dots}] plot coordinates{ (1,237.21)(2,145.46)(3,46.29)(4,23.18)}; \addlegendentry{\scriptsize BaDE \cite{kumar2020biphase}}
			\addplot+[ybar,fill=white, thick, pattern color=black!30, draw=orange!80!,  postaction={
				pattern=horizontal lines}] plot coordinates{ (1,223.41)(2,139.23)(3,41.18)(4,22.97)}; \addlegendentry{\scriptsize EQNN \cite{singh2021quantum}}
			\addplot+[ybar,fill=violet!10, thick, pattern color=violet!30, draw=violet!80!,  postaction={
				pattern=grid}] plot coordinates{ (1,54.14)(2,54.57)(3,31.11)(4,11.17)}; \addlegendentry{\scriptsize MCT-AQNN \cite{gupta2024multiple}}
			\addplot+[ybar,fill=blue!10!,thick, mark options={fill=white}, draw=blue, thick, pattern color=blue!100!, postaction={pattern= crosshatch dots, thick}] plot coordinates{ (1,53.82)(2,43.53)(3,14.66)(4,7.38)}; \addlegendentry{\scriptsize CA-QNN}
		\end{axis}
	\end{tikzpicture}
	\caption{Training time elapsed} \label{fig:training time}
\end{subfigure}
\hfill
\begin{subfigure}[t]{0.49\textwidth}
	\centering
	\begin{tikzpicture}
		\begin{axis}[
			width=.99\columnwidth,
			height=6.0cm,
			ymin=0,
			ymax=60,ytick={0,10,20,30,40,50,60},yticklabel style={font={\scriptsize}},
			xtick={1,2,3,4}, xticklabels={5,10,30,60},xticklabel style={font={\scriptsize}},
			ybar,
			ymajorgrids=true,
			xmajorgrids=true,
			grid style=dashdotted,
			bar width=6.0pt,
			enlarge x limits=0.12,
			xlabel={\textit{ Prediction Interval (minutes)}},
			nodes near coords, 
			every node near coord/.append style={rotate=90, anchor=west, color=black,font={\scriptsize}},
			legend style={at={(0.5,1.2)},
				anchor=north,legend columns=5},
			ylabel={\textit{Epochs}}]
			\addplot+[ybar, fill=white, thick, draw=teal, pattern color=black!30, postaction={
				pattern=grid,thin}] plot coordinates{ (1,39)(2,51)(3,26)(4,26)}; \addlegendentry{\scriptsize SaDE \cite{kumar2018workload}}
			
			\addplot+[ybar,fill=white,thick, pattern color=black!30, draw=olive, postaction={
				pattern=dots}] plot coordinates{ (1,23)(2,24)(3,24)(4,23)}; \addlegendentry{\scriptsize BaDE \cite{kumar2020biphase}}
			
			\addplot+[ybar,,fill=white, thick, pattern color=black!30, draw=orange!80!,  postaction={
				pattern=horizontal lines}] plot coordinates{ (1,21)(2,20)(3,21)(4,22)}; \addlegendentry{\scriptsize EQNN \cite{singh2021quantum}}
			
			\addplot+[ybar,fill=violet!10, thick, pattern color=violet!30, draw=violet!80!,  postaction={
				pattern=grid}] plot coordinates{(1,15)(2,17)(3,19)(4,21)}; \addlegendentry{\scriptsize MCT-AQNN \cite{gupta2024multiple}}
			
			\addplot+[ybar,fill=blue!10!,thick, mark options={fill=white}, draw=blue, thick, pattern color=blue!100!, postaction={pattern= crosshatch dots, thick}] plot coordinates{ (1,15)(2,16)(3,18)(4,20)}; \addlegendentry{\scriptsize CA-QNN}
		\end{axis}
	\end{tikzpicture}
	\caption{Number of epochs }
\end{subfigure}
\caption{Comparison of prediction models optimization}
\label{fig:comp_time_epochs}
\end{figure*}
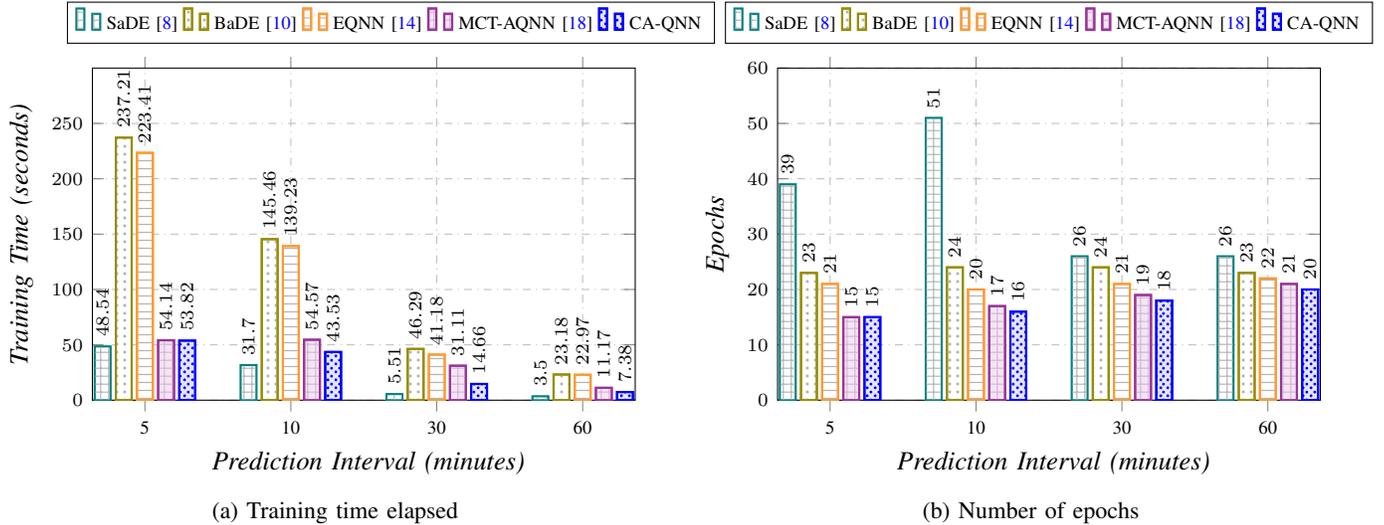

The training time consumed for the models is compared in Fig. \ref{fig:comp_time_epochs} (a), which decreases in the order of BaDE $>$ EQNN $>$ MCT-AQNN $>$ \textbf{CA-QNN} $>$ SaDE  with respect to GCD-CPU for PI of five minutes. The training time of CA-QNN is moderate compared with state-of-the-art models due to its dynamic multi-dimensional architectural optimization capabilities which provisions an opportunity to learn and upgrade faster, while the other models have static optimization which consumes longer time.  Fig. \ref{fig:comp_time_epochs} (b) presents the number of epochs elapsed in descending order as  SaDE $>$ BaDE $>$ EQNN $>$ MCT-AQNN $>$ \textbf{CA-QNN}. The proposed model  outperforms in terms of epochs elapsed due to faster convergence induced by variable length recombination along with higher learning  and strong control abilities.

Table \ref{tab:complexity} analyzes the complexities of CA-QNN and state-of-the-art models, wherein $n$, $m$, $t$, and $L$ representing number of input nodes, input samples, epochs elapsed, and size of population vectors, respectively. Time and space complexities decrease in the order of EQNN $\le$ SaDE $\equiv$ BaDE $\equiv$  MCT-AQNN $\equiv$  CA-QNN $>$ BPNN  for both time and space, attributed to the population-based optimization techniques used in these prediction models.
\begin{table}[!ht]
\caption{Computational complexity} \label{tab:complexity}
\centering
\resizebox{0.95\columnwidth}{!}{
	\tiny
	\begin{tabular}{lll}
		\hline
		\textbf{Model} &\textbf{Time Complexity} & \textbf{Space Complexity} \\ \hline
		BPNN \cite{lu2016rvlbpnn} &$\mathcal{O}(n^2mt)$ & $\mathcal{O}(n^2+mn)$  \\
		SaDE \cite{kumar2018workload} &$\mathcal{O}(n^2mtL)$  &$\mathcal{O}(n^2L+mn)$ \\
		BaDE \cite{kumar2020biphase} & $\mathcal{O}(n^2mtL)$ &$\mathcal{O}(n^2L+mn)$ \\ 
		EQNN \cite{singh2021quantum} & $\mathcal{O}(n^4tL)$ &$\mathcal{O}(n^2L+mn)$ \\
		MCT-AQNN \cite{gupta2024multiple} &$\mathcal{O}(n^2mtL)$  &$\mathcal{O}(n^2L+mn)$ \\
		\textbf{CA-QNN} &$\mathcal{O}(n^2mtL)$  &$\mathcal{O}(n^2L+mn)$ \\ \hline		
\end{tabular}}
\\  \footnotesize{\scriptsize $n$: \# Nodes in input layer, $m$: \# Input data samples, $t$: \# Generations/iterations, $L$: Size of VQNN Vector}
\end{table}

\subsection{ Ablation Study}
Table \ref{tab:ablation} presents the ablation study evaluating the impact of comprehensive learning based on both structural and parametric (qubit neural weights) optimization in our proposed CA-QNN model using different workload data traces, specifically GCD-CPU and GCD-Mem. The study contrasts three configurations: (i) \textit{both structural and parametric values varied}, (ii) \textit{fixed structure with varied parameters}, and (iii) \textit{both structure and parameters fixed}. The results clearly demonstrate that the fully adaptive configuration (varied-varied) significantly outperforms the other settings, achieving the lowest prediction errors across all metrics:  MAE, RMSE, and MAPE with values of 0.07245, 0.11866, and 0.38288 for GCPU and 0.06257, 0.09899, and 0.34450 for GMem, respectively.

\begin{table}[!htbp]
\caption{ Ablation Analysis: Effect of Structural Reconfiguration and Parametric Optimization on CA-QNN Performance} \label{tab:ablation}
\centering
\resizebox{0.95\columnwidth}{!}{
	\begin{tabular}{c|cc|ccc|c}
		\hline
		\multirow{2}{*}{\textbf{DT}} & \textbf{Structural} &\textbf{Parameter} & \multicolumn{3}{c|}{\textbf{Error Metrics}} & \textbf{Training } \\ 
		\cline{4-6}   &\textbf{Size} & \textbf{Values} &\textbf{MAE} &\textbf{RMSE} &\textbf{MAPE}& \textbf{Time (ms)} \\ \hline
		\multirow{3}{*}{\rotatebox{90}{GCPU}} &Varied &Varied & 0.07245 & 0.11866 & 0.38288 & 70031 \\ 
		\cline{2-7}&Fixed & Varied &0.18439 &0.17394  &0.43612 & 87776\\ 
		\cline{2-7}&Fixed & Fixed &0.36374 & 0.23487 &0.52231  & 95223\\ \hline \hline
		\multirow{3}{*}{\rotatebox{90}{GMem}} &Varied &Varied &0.06257 &0.09899 &0.34450 & 73036\\ 
		\cline{2-7}&Fixed & Varied &0.15311 &0.13212 &0.38742 & 81373 \\ 
		\cline{2-7}&Fixed & Fixed &0.28612 &0.18324 &0.41234 & 89273\\ \hline
\end{tabular}}
\\  \footnotesize{\scriptsize DT: Workload Data Traces, GCPU: GCD-CPU, GMem: GCD-Mem}
\end{table}
In contrast, the fixed-fixed configuration, which closely resembles traditional backpropagation-based neural network tuning, exhibits the highest error rates. This highlights its limited adaptability due to its inability to dynamically reconfigure both structure and parametric weights, resulting in suboptimal workload predictions. While allowing parametric variation (fixed-varied) provides some improvement over the rigid model, it still lags significantly behind the fully adaptive variant. This emphasizes the critical role of structural flexibility in capturing the complex, nonlinear dynamics of cloud workloads. Furthermore, despite the slight increase in computational time associated with adaptive structural optimization, the substantial reduction in prediction error justifies its adoption. The results validate the superiority of CA-QNN over conventional backpropagation-based neural networks by leveraging dynamic qubit-based neural weight adjustments and variable network configurations. This comprehensive learning strategy enables CA-QNN to achieve higher adaptability, scalability, and precision in cloud resource management, demonstrating its effectiveness in handling dynamic and unpredictable workload variations.

\section{Conclusion and Future work} \label{conclusion}
This work presented a novel CA-QNN prediction model, addressing the limitations of traditional neural networks in predicting diverse, high-dimensional cloud workloads. By integrating quantum computing with advanced structural and parametric learning-based qubit vector optimization, the CA-QNN model significantly enhances prediction accuracy, reducing errors by up to 93.40\% and 91.27\% compared to the existing methods. The extensive evaluation and comparison against seven state-of-the-art approaches across four benchmark datasets,  demonstrates superior performance of CA-QNN prediction model. This research paves the way for leveraging quantum mechanics in machine learning with complete network vector architectural optimization.   The future work will  aim to incorporate qubit-based transfer learning and enhance architectural optimization with multi-dataset learning, aiming to improve the generalization, accuracy, robustness, and adaptability of the model for unseen cloud workloads.
\bibliographystyle{IEEEtran}

\bibliography{bibfile}

\begin{IEEEbiography}[{\includegraphics[width=1.0in,height=1.25in,clip,keepaspectratio]{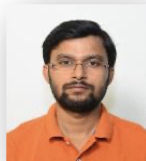}}]{Jitendra Kumar} (Senior Member, IEEE) is an Assistant Professor in the Department of Mathematics, Bioinformatics, and Computer Applications, Maulana Azad National Institute of Technology Bhopal, India. He obtained his PhD in Machine Learning and Cloud Computing from National Institute of Technology Kurukshetra, India in 2019. His current research interests include Cloud Computing, Computational Intelligence, Time Series Forecasting, Optimization, etc.  
\end{IEEEbiography}
\vspace{1pt}
\begin{IEEEbiography}[{\includegraphics[width=1.0in,height=1.25in,clip,keepaspectratio]{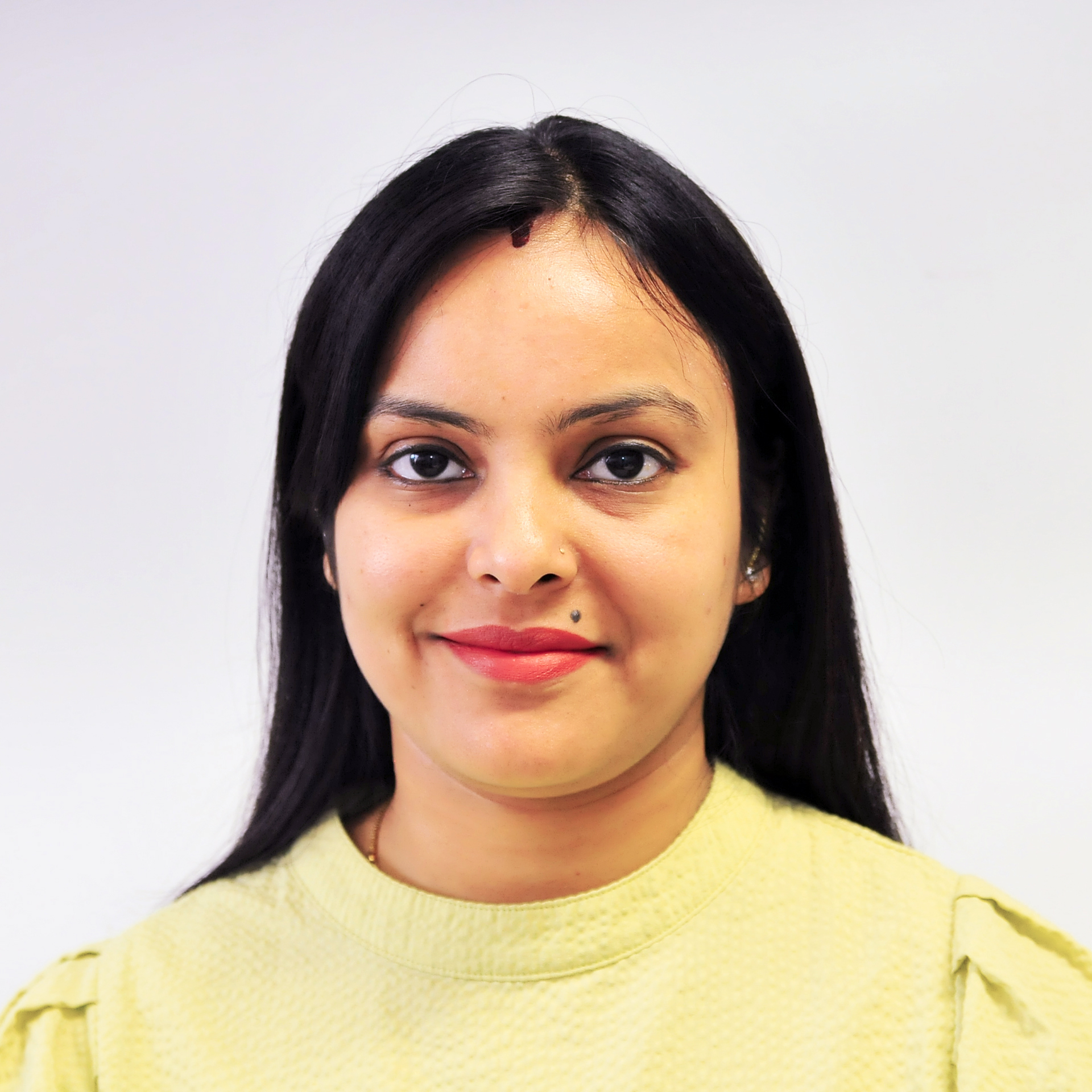}}]{Deepika Saxena} (Member, IEEE) is an Associate Professor at University of Aizu, Japan. She received her Ph.D. from National Institute of Technology Kurukshetra, India, and postdoctoral from Goethe University, Germany. She has received several prestigious awards, including IEEE TCSC Early Career Researcher Award 2024, IEEE TCSC Outstanding Ph.D. Dissertation Award 2023, EUROSIM Best Ph.D. Thesis Award 2023, and IEEE Computer Society Best Paper Award 2022. She is also a recipient of the JSPS KAKENHI Early Career Young Scientist Research Grant FY2024. Her research focuses on Neural Networks, Evolutionary Algorithms, Cloud Computing, Digital Twins, Quantum Machine Learning, Cyber-Security Intelligence, and Vehicle Traffic Management.  
\end{IEEEbiography}
\vspace{1pt}
\begin{IEEEbiography}[{\includegraphics[width=1.0in,height=1.25in,clip,keepaspectratio]{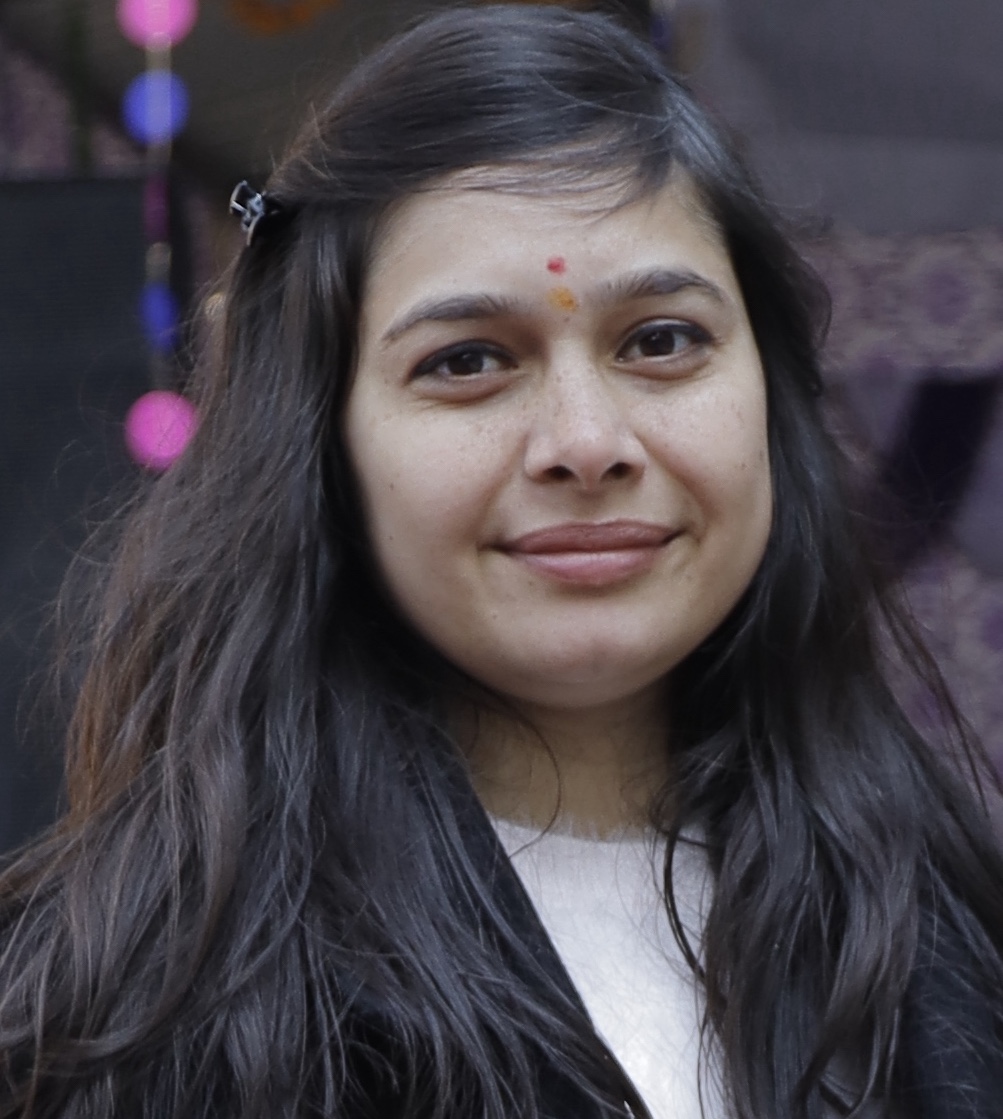}}]{Kishu Gupta} (Member, IEEE) is working as a Post Doctoral Researcher at Department of Computer Science and Engineering, National Sun Yat-sen University, Kaohsiung, Taiwan.  
She received Ph.D. from India in 2023, receiving INSPIRE Fellowship. Also, she is honored a GOLD MEDAL for securing First Rank in her post-graduation and BEST PAPER Award in an International Conference RTIP2R-2024 organized with India-USA collaboration. Her major research interests include data security and privacy, cloud computing, FedL, and quantum ML. 
\end{IEEEbiography}
\vspace{1pt}
\begin{IEEEbiography}[{\includegraphics[width=1in,height=1.25in,clip,keepaspectratio]{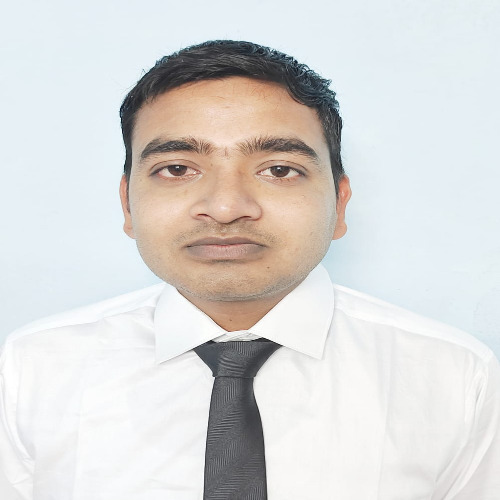}}]{Satyam Kumar} is currently working with Carelon Global Solutions. He received his masters degree (M.Tech.) from the Department of Computer Applications, National Institute of Technology Tiruchirappalli, India. His research interests are Cloud Computing, Machine Learning, and Optimisation.
\end{IEEEbiography}
\vspace{1pt}
\begin{IEEEbiography}[{\includegraphics[width=1.0in,height=1.25in,clip,keepaspectratio]{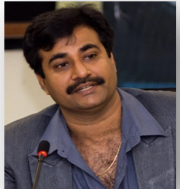}}]{Ashutosh Kumar Singh} (Senior Member, IEEE) is working as a Professor and Director of Indian Institute of Information Technology Bhopal, India. Also, he is working as Adjunct Professor in the VIZJA University, Warsaw, Poland. He received his Ph.D. in Electronics Engineering from Indian Institute of Technology, BHU, India and Post Doc from Department of Computer Science, University of Bristol, UK. His research area includes Design and Testing of Digital Circuits, Data Science, Cloud Computing, Machine Learning, Security. He has published more than 400 research papers in different journals and conferences of high repute. He received IEEE Computer Society Best Paper Award 2022.
\end{IEEEbiography}

\end{document}